\documentclass[runningheads]{llncs}

\usepackage{eccv}

\usepackage{eccvabbrv}

\usepackage{graphicx}
\usepackage{booktabs}
\usepackage{multirow, array}  % new add
\usepackage{endnotes}  % new add
\usepackage{threeparttable} % new add
\usepackage{colortbl}
\definecolor{shade}{RGB}{127,127,127} 
\usepackage{authblk}

\usepackage[accsupp]{axessibility}  % Improves PDF readability for those with disabilities.

\usepackage{hyperref}

\usepackage{orcidlink}

\begin{document}

\title{SHINE: \underline{S}aliency-aware \underline{HI}erarchical \underline{NE}gative Ranking for Compositional Temporal Grounding} 

\titlerunning{Coarse-to-Fine Saliency Ranking for Compositional Temporal Grounding}

\author{Zixu Cheng$^\star$\inst{1}  \and Yujiang Pu\thanks{Equal Contribution}\inst{2} \and Shaogang Gong\inst{1} \and  Parisa Kordjamshidi\inst{2} \and Yu Kong\inst{2}}

\authorrunning{Cheng and Pu et al.}

\institute{
Queen Mary University of London, London, UK\\ \email{\{zixu.cheng, s.gong\}@qmul.ac.uk}  \and
Michigan State University, East Lansing, US\\ \email{\{puyujian, kordjams, yukong\}@msu.edu}
}

\maketitle

\begin{abstract}
Temporal grounding, also known as video moment retrieval, aims at locating video segments corresponding to a given query sentence. The compositional nature of natural language enables the localization beyond predefined events, posing a certain challenge to the compositional generalizability of existing methods. Recent studies establish the correspondence between videos and queries through a decompose-reconstruct manner to achieve compositional generalization. However, they only consider dominant primitives and build negative queries through random sampling and recombination, resulting in semantically implausible negatives that hinder the models from learning rational compositions. In addition, recent DETR-based methods still underperform in compositional temporal grounding, showing irrational saliency responses when given negative queries that have subtle differences from positive queries. To address these limitations, we first propose a large language model-driven method for negative query construction, utilizing GPT-3.5 Turbo to generate semantically plausible hard negative queries. Subsequently, we introduce a coarse-to-fine saliency ranking strategy, which encourages the model to learn the multi-granularity semantic relationships between videos and hierarchical negative queries to boost compositional generalization. Extensive experiments on two challenging benchmarks validate the effectiveness and generalizability of our proposed method. Our code is available at \href{https://github.com/zxccade/SHINE}{https://github.com/zxccade/SHINE}.

  \keywords{Temporal Grounding \and Compositional Generalization}
\end{abstract}

\section{Introduction}
\label{sec:intro}

Temporal grounding \cite{gao2017tall, krishna2017dense, zhang2020span, moon2023query} has received continuous attention for its wide range of applications in video summarization~\cite{li2023progressive, he2023align}, moment retrieval~\cite{zala2023hierarchical, lin2023univtg}, surveillance and security~\cite{pu2022locality, pu2023learning}. Unlike typical temporal action localization tasks \cite{zhu2021enriching}, temporal grounding aims to retrieve video moments based on textual queries that include not only the action itself but also the objects, attributes, and interactions.
Moreover, the extensive vocabulary of natural language can expand the limited terms from labeled videos to describe new, unlabeled scenarios in the training data. However, recent works \cite{yuksekgonul2022and, doveh2023dense, trager2023linear, singh2023coarse} have shown that existing vision-language models (VLMs) lack compositional generalizability for unseen combinations, as reflected by their insensitivity to word order and primitives.

To address this challenge, compositional temporal grounding has been recently proposed in \cite{li2022compositional}, which aims at locating unseen video moments by learning novel combinations of known words in the training data. They decompose videos and queries into global events, local actions, and atomic objects, and establish relationships between visual and text concepts by constructing a hierarchical semantic graph. However, their pipeline greatly relies on off-the-shelf object detection and action recognition models, which exhibits poor flexibility and scalability. Later, Li \etal \cite{li2023exploring} proposed a self-supervised learning framework to enhance the compositional generalization capability of existing VLMs by masking different primitives to generate semantically equivariant and invariant samples. In contrast, Deco \cite{yang2023deco} has constructed negative samples through a decompose-reconstruct strategy, employing a mask-and-predict ranking loss to learn the multi-granularity correspondence between video-text pairs.

While these methods have improved existing techniques, they share limitations in the construction of negative samples: (1) VISA \cite{li2022compositional} and SSL2CG \cite{li2023exploring} focus on dominant verbs and nouns, but overlook the role of other primitives such as prepositions and adverbs (\eg, \emph{on}/\emph{under the table} and \emph{turn on}/\emph{off the light}), where the substitution of these words fundamentally changes the semantics. (2) Deco \cite{yang2023deco} neglects the viability of semantics when recombining primitives, resulting in numerous infeasible combinations, such as \emph{eating the table} and \emph{reading the door}. These issues not only hinder the model from learning the semantics of non-dominant primitives, but also force the model to extract differences from unrealistic combinations.

\begin{figure}[tb]
  \centering
  \setlength{\belowcaptionskip}{-10pt}
  \includegraphics[width=0.9\textwidth]{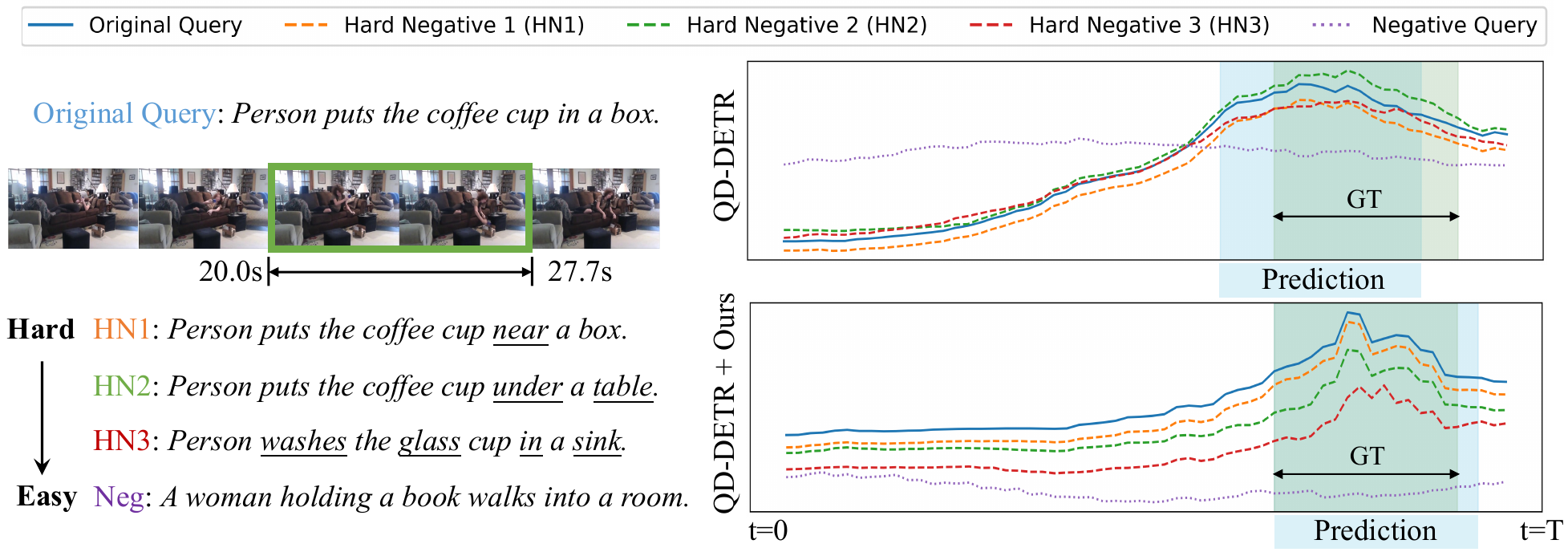}
  \caption{Comparison of saliency scores given different queries. The existing work \cite{moon2023query} struggles with discerning hard negative queries, showing irrational saliency responses under different primitive substitutions. Our method helps a model to learn the nuances in the semantics of hierarchical negative samples, suppressing the model's response to irrelevant queries while boosting its compositional generalizability.
  }
  \label{fig:motivation}
\end{figure}

Additionally, these works \cite{li2022compositional, li2023exploring, yang2023deco} have only explored the compositional generalizability of classic temporal grounding methods, lacking considerations on recent novel architectures, such as DETR-based methods \cite{lei2021detecting, liu2022umt, moon2023query, jang2023knowing}. These approaches combine highlight detection \cite{yao2016highlight} with the temporal grounding task, aiming to locate segments corresponding to the query while predicting the saliency scores for each moment. The saliency scores evaluate the relevance of all video clips to a given query, revealing the corresponding highlight moments. However, we observed that existing work \cite{moon2023query} struggles with discerning different negative queries, showing irrational saliency responses under different primitive substitutions, as shown in \cref{fig:motivation}. This indicates that current approaches tend to ignore the nuances between hard negative and positive queries. Consequently, they fail to accurately match visual representations with corresponding primitives, hindering their ability to achieve compositional generalization.

To this end, we first propose a large language model (LLM)-driven approach for constructing hard negative samples, which are semantically plausible yet distinct from the original query. With these manipulated negatives at hand, we further introduce a coarse-to-fine saliency ranking strategy to establish a multi-granularity semantic relationship between video clips and hierarchical negative queries. Compared to existing works, our method has the following advantages: (1) a good compositional representation for negative queries that consider the significance of different primitives while maintaining semantic feasibility; (2) the saliency scores derived from negative samples at different levels exhibit a hierarchical divergence, indicating that our method successfully captures the multi-granularity relationship between video clips and queries; (3) our method can be seamlessly integrated into existing DETR-based models, significantly improving their generalization capabilities to unseen combinations while maintaining the accuracy for seen samples. In summary, our contributions are three-fold: 
\begin{itemize}
    \item To address the issue of implausible negative queries generated by random sampling, we introduce an LLM-driven approach that produces semantically viable hard negative queries, which facilitates temporal grounding models to learn plausible compositional semantics.
    \item To deal with the irrational saliency responses in existing methods, we propose a coarse-to-fine saliency ranking strategy that utilizes the plausible hard negatives to capture hierarchical semantic differences and boost their compositional generalizability.
    \item Extensive experiments are conducted with two DETR-based backbones on two challenging benchmarks, Charades-CG and ActivityNet-CG, which show that our method significantly improves baseline performance and achieves competitive results.
\end{itemize}

\section{Related Work}
\textbf{Temporal Grounding.} Temporal grounding, \emph{a.k.a.} video moment retrieval, initially proposed in \cite{gao2017tall} and \cite{krishna2017dense}, aims at localizing segments in a video that match the description of a query sentence. Currently, dominant supervised learning techniques are divided into two categories: proposal-based and proposal-free methods. Proposal-based methods generate candidate segments through various strategies, including sliding windows\cite{gao2017tall, liu2018cross}, dense proposals\cite{xu2019multilevel, xiao2021boundary, hou2022cone}, and fixed anchors\cite{chen2018temporally, zhang2019man, zhang2020learning, wang2022negative}, subsequently selecting the most appropriate intervals based on a similarity measure. However, the generation of candidate proposals and their semantic matching with queries are computationally intensive. Conversely, proposal-free methods \cite{yuan2019find, mun2020local, zeng2021multi, yan2023unloc, lin2023univtg} directly predict the temporal boundaries of the target clip. This paradigm eliminates the need for proposal generation, significantly enhancing the model's efficiency during inference. Recently, Lei \etal \cite{lei2021detecting} reformulated the temporal grounding task as a set prediction problem, introducing a DETR-based \cite{carion2020end} architecture enabling simultaneous video moment retrieval and highlight detection. Subsequent works, including UMT \cite{liu2022umt}, QD-DETR \cite{moon2023query}, and EaTR \cite{jang2023knowing}, have enhanced localization accuracy by refining the DETR framework. Differently, our work establishes a connection between the saliency score and compositionality, which effectively unlocks the potential for compositional generalization in DETR-based models.

\noindent\textbf{Compositional Generalization.} Recently, the compositional generalizability of vision-language models, or VLMs, has received sustained attention \cite{yuksekgonul2022and, doveh2023dense, trager2023linear, xu2023metarevision, xu2024gipcol}, with several benchmarks being proposed for evaluating the robustness of the models on specific downstream tasks, including image-text retrieval \cite{ma2023crepe, ray2023cola, hsieh2023sugarcrepe}, visual question answering \cite{johnson2017clevr, grunde2021agqa, gandhi2022measuring, yu2023anetqa}, and zero-shot learning \cite{li2022siamese, lu2023decomposed, zheng2024caila}. To further evaluate the compositional generalization of existing temporal grounding methods, Li \etal \cite{li2022compositional} has constructed two benchmarks, Charades-CG and ActivityNet-CG, and proposed a variational cross-graph reasoning framework to achieve compositional video-text comprehension. Later, Li \etal \cite{li2023exploring} generated semantic equivariant and invariant samples by masking different primitives, and employed contrastive learning to improve the compositional generalization capability of existing VLMs. Yang \etal \cite{yang2023deco} constructed negative queries through a decompose-reconstruct strategy, utilizing a mask-and-predict contrastive ranking loss to learn the multi-granularity correspondence between video-text pairs. However, most of these works only consider dominant primitives like verbs and nouns, ignoring the effect of other words like prepositions and adverbs. Moreover, they adopt random sampling to replace primitives for negative construction, which hinders the model from learning semantically feasible compositions. In contrast, we progressively replace the primitives with different ratios beyond just verbs and nouns. Additionally, we resort to a large language model to generate semantically plausible negative queries instead of random sampling.

\section{Method}

\begin{figure}[tb]
  \centering
  \setlength{\belowcaptionskip}{-10pt}
  \includegraphics[width=\textwidth]{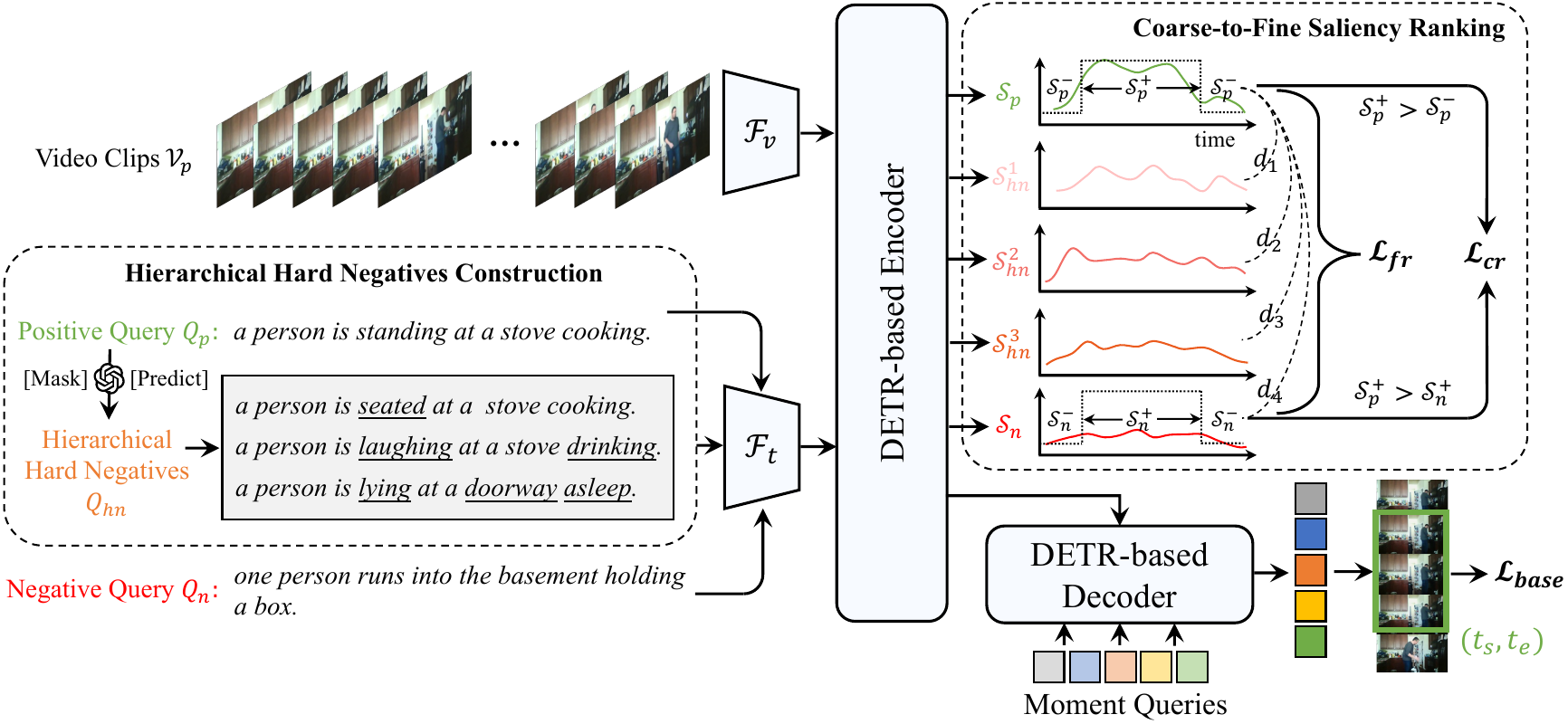}
  \caption{The overall framework of our method SHINE. For each video-text pair, we first generate a set of hierarchical hard negative queries and randomly sample one negative query from the same mini-batch. These queries and the video clips are fed into a DETR-based encoder for interaction and predicting saliency scores $S$. The coarse-grained ranking loss $\mathcal{L}_{cr}$ aims to enlarge the disparity between the saliency scores produced by positive and negative queries, and the fine-grained ranking loss $\mathcal{L}_{fr}$ is designed to capture the nuanced semantics among the hierarchical hard negative queries. These two constraints are combined with $\mathcal{L}_{base}$ to optimize the model.
  }
  \label{fig:framework}
\end{figure}

\subsection{Problem Definition and Overview}
Given an untrimmed video $V$ and a query sentence $Q$, our goal is to identify the start and end timestamps $(t_s, t_e)$ of the moments in the video that correspond to the query. The model is expected to achieve precise localization based on the novel combinations of seen words in the training set. Compared to the conventional temporal grounding task, we seek a good balance in performance between seen and unseen compositions, which requires the model to avoid overfitting while ensuring compositional generalizability.

% \com{this paragraph is way too long}
The overall framework of our proposed method is shown in \cref{fig:framework}. Given a video-query pair $(V_p,Q_p)$, we first construct a set of hierarchical hard negative queries $\{Q^i_{hn}\}^3_{i=1}$ via a progressive mask-and-predict strategy (\cref{subsec:hhn}). Notably, we utilize a large language model, namely GPT-3.5 Turbo \cite{brown2020language}, to select appropriate words from the training set for primitive replacement. This operation gradually changes the semantics of the original query while effectively avoiding implausible combinations. These manipulated queries, along with the original query $Q_p$ and a negative query $Q_n$ selected from other queries in the same mini-batch, constitute a set of queries. During the training phase, each video and its associated query set are first fed into a video encoder $\mathcal{F}_v(\cdot)$ and a text encoder $\mathcal{F}_t(\cdot)$ to extract corresponding features, which are then processed by the encoder to predict the saliency scores $\{S_p, S^1_{hn}, S^2_{hn}, S^3_{hn}, S_n\}$. Subsequently, we propose modeling the video-level saliency prior using a coarse-grained saliency ranking loss $\mathcal{L}_{cr}$, which incorporates two constraints designed to enhance the discriminability between positive and negative queries (\cref{subsec:cgssr}). Concurrently, we employ a fine-grained saliency ranking loss $\mathcal{L}_{fr}$ to discern the saliency scores derived from the query set, which facilitates the learning of multi-granularity semantics by exploring the nuance among hierarchical negative samples. By combining $\mathcal{L}_{base}$ with the proposed coarse-to-fine saliency ranking loss (\cref{subsec:loss}), our method can be seamlessly integrated into existing DETR-based models \cite{lei2021detecting, moon2023query}, greatly enhancing their potential for compositional generalization.

\subsection{Hierarchical Hard Negatives Construction}
\label{subsec:hhn}

Unlike previous methods that only consider verbs and nouns for negative query construction, we argue that other primitives like adverbs and prepositions also play an important role in composition generalization. Moreover, negative samples generated by randomly replacing primitives contain a large number of semantically infeasible combinations. Accordingly, we first propose an LLM-driven method to construct semantically viable hierarchical negative queries.

\begin{figure}[tb]
  \centering
  \setlength{\belowcaptionskip}{-10pt}
  \includegraphics[width=0.9\textwidth]{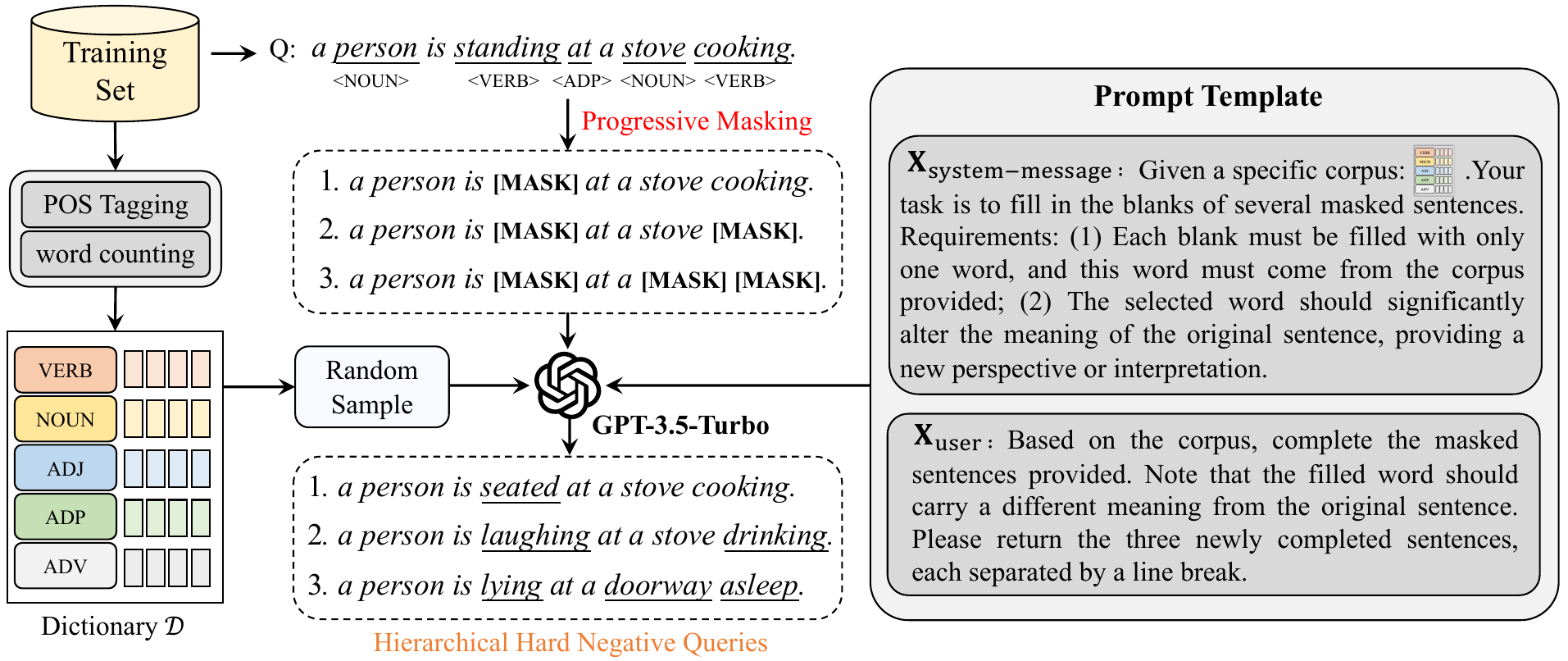}
  \caption{The construction pipeline of hierarchical hard negative queries.
  }
  \label{fig:negative}
\end{figure}

\cref{fig:negative} shows the pipeline for constructing hierarchical negative queries. Specifically, we first use spaCy\footnote{spaCy: https://spacy.io/} to perform part-of-speech (POS) tagging on all query sentences in the training set, and construct a dictionary $D$ by counting words from five types of primitives (verbs, nouns, adjectives, prepositions, and adverbs). Subsequently, for a given query $Q$, we progressively mask the primitives with different ratios in the original query according to their relative importance in linguistics, \ie, verb-noun-adjective-preposition-adverb. Also, to ensure contextual consistency in negative queries, subjects (usually nouns) are only considered when other primitives are insufficient.
% Following this, for a given query $Q$ and its corresponding word tags, we adopt a progressive mask-and-predict strategy that considers the significance of the primitive words to create fill-in-the-blank templates of varying difficulty. 
Instead of filling in the masks with random selections from $D$, we resort to a powerful LLM, \ie, GPT-3.5 Turbo, to generate semantically plausible hard negative queries. Considering the token limit inherent to LLMs, we randomly select a subset from the dictionary to ensure a balance between context and diversity for each sample. Furthermore, we have carefully crafted a prompt template to steer the LLM toward producing challenging negative queries. These negatives are semantically viable yet distinct from the original query, laying the foundation for fine-grained saliency ranking.

\subsection{Coarse-to-Fine Saliency Ranking}
\label{subsec:cgssr}

The saliency scores measure the relevance of video clips to a given query, revealing the corresponding highlight moments. However, we observed that existing methods \cite{lei2021detecting, moon2023query} exhibit irrational saliency responses when faced with different negative queries, as shown in \cref{fig:motivation}. The saliency scores of some hard negative queries even surpass that of the original ones. This indicates that current methods fail to accurately match visual representations with corresponding primitives, struggling to discern the nuances between different queries. To this end, we introduce a coarse-to-fine saliency ranking strategy by establishing a multi-granularity semantic relationship between video clips and different negative queries. This approach enables the model to capture hierarchical semantic differences and enhances its compositional generalizability.

\begin{figure}[tb]
  \centering
  \setlength{\belowcaptionskip}{-10pt}
  \includegraphics[width=0.9\textwidth]{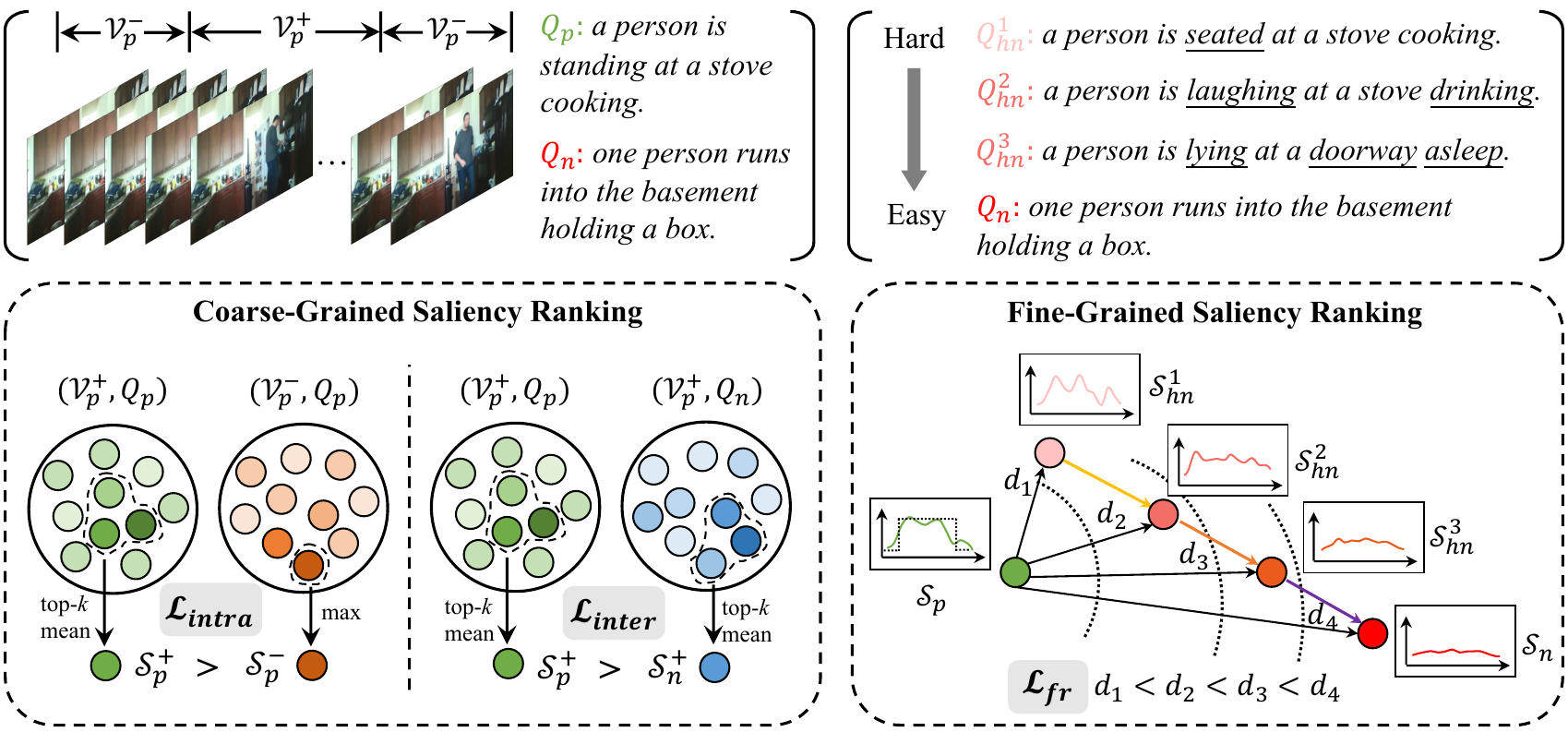}
  \caption{An illustration of the Coarse-to-Fine Saliency Ranking strategy.
  }
  \label{fig:CFSR}
\end{figure}

\noindent \textbf{Coarse-Grained Saliency Ranking.} As shown in \cref{fig:CFSR}, for a given video-text pair $(V_p, Q_p)$, we designate video clips within and outside the ground-truth interval as $V_p^+$ and $V_p^-$, respectively, and $Q_p$ as the corresponding positive query. Concurrently, a negative query is randomly selected from the same mini-batch, denoted as $Q_n$. 

Intuitively, the saliency scores within the ground-truth interval should surpass those outside, and scores elicited by positive queries should exceed those by negative ones. With these two priors, we introduce a coarse-grained ranking loss with dual constraints, formulated as:

\begin{equation}
  \mathcal{L}_{cr}=\underbrace{\max(0, h_1+S^-_p - S^+_p)}_{\mathcal{L}_{intra}} + \underbrace{\max(0, h_2+S^+_n - S^+_p)}_{\mathcal{L}_{inter}},
  \label{eq:coarse}
\end{equation}
\begin{equation}
  S^+=\frac{1}{k}\sum^k_{i=1} \text{sort}(S)_{T^+},  k = \max(1, \lfloor T^+/q \rfloor),
  \label{eq:topk}
\end{equation}
where $S^+_p$ and $S^+_n$ represent the top-\(k\) mean value of saliency scores yielded by $(V^+_p, Q_p)$ and $(V^+_p, Q_n)$, respectively, and $S^-_p$ is the maximum of saliency scores produced by $(V^-_p, Q_p)$. $T^+$ denotes the number of clips within the ground-truth interval, and $q$ is a factor to control the selection ratio. $h_1$ and $h_2$ are two predefined margins. Notably, our constraints differ from \cite{lei2021detecting} in two key ways: (1) Rather than using a single maximum value, we adapt to intervals of different scales based on interval length and $q$. (2) Beyond considering the internal difference of positive queries, we also enlarge the saliency gap between positive and negative queries to achieve better discriminability. 

\noindent \textbf{Fine-Grained Saliency Ranking.} While the coarse-grained ranking loss improves the discriminative capability of the video-text representation, it does not fully capture the relationships between the query primitives and the video clip. In DETR-based architectures, saliency scores are temporally aligned with the timestamps of localization boundaries, mirroring the relevance of the current video clip to the query. We argue that saliency scores tied to the original query should be temporally consistent with the ground truth, whereas those related to hard negatives ought to display a hierarchical disparity from the positive one, as illustrated in \cref{fig:CFSR}. Based on this assumption, we further propose a fine-grained ranking loss to refine the saliency scores derived from varying semantic levels of negative queries, formulated as:

\begin{equation}
\begin{aligned}
\mathcal{L}_{fr} = & \max(0, m_0 + d(Y, S_p) - d(S_p, S^1_{hn})) \\
& + \max(0, m_1 + d(S_p, S^1_{hn}) - d(S_p, S^2_{hn})) \\
& + \max(0, m_2 + d(S_p, S^2_{hn}) - d(S_p, S^3_{hn})) \\
& + \max(0, m_3 + d(S_p, S^3_{hn}) - d(S_p, S_{n})),
\end{aligned}
\label{eq:fine}
\end{equation}

\begin{equation}
d(y, \hat{y}) = -\frac{1}{T}\sum_{i=1}^T y_i\log(\hat{y}_i),
\label{eq:neg-log-likelihood}
\end{equation}
where $\{S^i_{hn}\}^3_{i=1}$ denotes saliency scores of hierarchical negative queries,  $d(\cdot)$ symbolizes the negative log-likelihood between the observation \(y\) and the prediction \(\hat{y}\), measuring the disparity in the distribution of saliency scores across the temporal dimension $T$. In particular, due to the lack of ground-truths for the saliency scores, a value of 1 is assigned to moments inside the localization interval, and 0 is assigned to moments outside the interval, with $Y$ denoting the pseudo saliency score. \(m_0\) to \(m_3\) represent four predefined margins. This loss not only underscores the nuanced semantics between the video and the query sentence, but also the differences in temporal distribution and magnitude of the saliency scores in the hard negative samples. By hierarchically constraining the saliency scores of these hard negatives, our method helps the model discern the nuances between various primitive words and video moments, suppressing irrational saliency responses, and further improving its capability to identify novel combinations.

\subsection{Model Training Objectives}
\label{subsec:loss}
Since our method can be seamlessly integrated into existing DETR-based models, during the training phase, we optimize the model utilizing three distinct loss functions, with the overall objective expressed as:
\begin{equation}
\mathcal{L} = \mathcal{L}_{base} + \alpha\mathcal{L}_{cr} + \beta\mathcal{L}_{fr}
\label{eq:overall}
\end{equation}
where $\mathcal{L}_{base}$ represents the basic loss of the DETR-based model, typically including bipartite matching loss, moment localization loss, and saliency loss. More details can be found in \cref{appendix}. $\alpha$ and $\beta$ are two weight coefficients. By combining the two constraints with $\mathcal{L}_{base}$, our method can significantly enhance the model's compositional generalization capabilities while preserving accuracy for in-distribution samples, as demonstrated in subsequent experiments.

\section{Experiments}

% \subsection{Experimental Settings}
\noindent \textbf{Datasets and Evaluation Metrics.}
We evaluate our method on two newly proposed benchmarks, Charades-CG and ActivityNet-CG\cite{li2022compositional}, originated from Charades-STA \cite{gao2017tall} and ActivityNet Captions \cite{krishna2017dense}. Each dataset is reorganized into four splits: Training/Test-Trivial/Novel-Composition/Novel-Word, where the latter three splits evaluate the model's performance on IID samples, novel combinations of seen words, and unseen words, respectively. In particular, the Novel-Composition split considers five types of new compositions: verb-noun, noun-noun, verb-adverb, adjective-noun, and preposition-noun. Following previous works \cite{li2022compositional, li2023exploring, yang2023deco}, we use two main metrics to evaluate our methods, \ie, "$\text{R}@\text{n}, \text{IoU}=m$" and mean Intersection over Union (mIoU). The former denotes the percentage of queries with at least one prediction whose IoU score is larger than $m$ within the top-$n$ predictions, while the latter refers to the average of IoU scores across all queries.

\noindent \textbf{Implementation Details.} 
We adopt Moment-DETR\cite{lei2021detecting} and QD-DETR\cite{moon2023query} as our baselines and integrate them with our method using their officially released code. Unless specifically noted, other hyperparameters follow their default settings.
For QD-DETR, following \cite{yang2023deco, li2023exploring}, we utilize pretrained I3D \cite{carreira2017quo} features on the Charades-CG dataset and C3D \cite{karpathy2014large} features on the ActivityNet-CG dataset, respectively. Following \cite{lei2021detecting}, we employ SlowFast \cite{feichtenhofer2019slowfast} and CLIP \cite{radford2021learning} to extract hybrid visual features for Moment-DETR.
As for text features, we follow \cite{lei2021detecting, moon2023query} to extract CLIP features with 512 dimensions for each query. 
In hierarchical hard negative construction, we progressively mask the original query in Charades-CG at ratios of 25\%, 50\%, and 75\%, while the masking ratios for ActivityNet-CG are set to 10\%, 30\%, and 50\%. For Charades-CG, we set the learning rates for QD-DETR and Moment-DETR to 0.0001 and 0.0002, respectively, while for ActivityNet-CG, the learning rates for both models are set to 0.0002. The coarse-grained margins $h_1$ and $h_2$ are set to 1.0 and 2.0, respectively, while the relative thresholds $m_0$ to $m_3$ are set to 0.25. The factor $q$ in \cref{eq:topk} are set to 8 for both Charades-CG and ActivityNet-CG. All experiments are run on a single NVIDIA A100 GPU with a batch size of 32 training for 200 epochs.

\subsection{Comparisons with the State-of-the-arts}

\renewcommand{\arraystretch}{1.2}
\begin{table}[tb]
\caption{Performance (\%) of state-of-the-art methods on the Charades-CG dataset. The best result is shown in \textbf{bold} and the second best is \underline{underlined}. `WS': weakly-supervised methods. `RL': reinforcement learning methods. `PB': proposal-based methods. `PF': proposal-free methods. $\dag$ indicates the results of our implementation using the officially released code. \textcolor{shade}{$\star$ denotes the results relying on external detector knowledge.}}
\label{tab:charades}
\centering
\scalebox{0.7}{% Start of \resizebox
\begin{tabular}{llccccccccc}
\toprule
\multirow{2}{*}{Setting} & \multirow{2}{*}{Method} & \multicolumn{3}{c}{Test-Trivial} & \multicolumn{3}{c}{Novel-Composition} & \multicolumn{3}{c}{Novel-Word} \\ \cmidrule(lr){3-5} \cmidrule(lr){6-8} \cmidrule(l){9-11}
 & & R1@0.5 & R1@0.7 & mIoU & R1@0.5 & R1@0.7 & mIoU & R1@0.5 & R1@0.7 & mIoU \\  \midrule
WS & WSSL \cite{duan2018weakly} & 15.33 & 5.46 & 18.31 & 3.61 & 1.21 & 8.26 & 2.79 & 0.73 & 7.92 \\  \midrule
RL & TSP-PRL \cite{wu2020tree} & 39.86 & 21.07 & 38.41 & 16.3 & 2.04 & 13.52 & 14.83 & 2.61 & 14.03 \\ \hline
\multirow{3}{*}{PB} & TMN \cite{liu2018temporal} & 18.75 & 8.16 & 19.82 & 8.68 & 4.07 & 10.14 & 9.43 & 4.96 & 11.23 \\
 & 2D-TAN \cite{zhang2020learning} & 48.06 & 27.10 & 43.72 & 32.74 & 15.25 & 31.5 & 37.12 & 18.99 & 35.04 \\
 & 2D-TAN+SSL \cite{li2023exploring} & 53.91 & 31.82 & 46.84 & 35.42 & 17.95 & 33.07 & 43.60 & 25.32 & 39.32 \\ 
 & MS-2D-TAN \cite{zhang2021multi} & 57.85 & 37.63 & 50.51 & 43.17 & 23.27 & 38.06 & 45.76 & 27.19 & 40.80 \\
 & MS-2D-TAN+SSL \cite{li2023exploring} & 58.14 & \underline{37.98} & 50.58 & 46.54 & \underline{25.10} & 40.00 & \underline{50.36} & \underline{28.78} & \underline{43.15} \\  \midrule
\multirow{6}{*}{PF} & LGI \cite{mun2020local} & 49.45 & 23.8 & 45.01 & 29.42 & 12.73 & 30.09 & 26.48 & 12.47 & 27.62 \\
 & VLSNet \cite{zhang2020span} & 45.91 & 19.80 & 41.63 & 24.25 & 11.54 & 31.43 & 25.60 & 10.07 & 30.21 \\
 & \textcolor{shade}{$\text{VISA}^\star$} \cite{li2022compositional} & \textcolor{shade}{53.20} & \textcolor{shade}{26.52} & \textcolor{shade}{47.11} & \textcolor{shade}{45.41} & \textcolor{shade}{22.71} & \textcolor{shade}{42.03} & \textcolor{shade}{42.35} & \textcolor{shade}{20.88} & \textcolor{shade}{40.18} \\
 & Deco \cite{yang2023deco} & 58.75 & 28.71 & 49.06 & \underline{47.39} & 21.06 & \underline{40.70} & - & - & - \\ \cmidrule{2-11}
 
 & $\text{Moment-DETR}^\dag$ \cite{lei2021detecting} & 49.48 & 28.04 & 44.82 & 39.42 & 18.62 & 36.61 & 46.76 & 24.75 & 41.70  \\ 
 & \cellcolor{gray!25}Moment-DETR+Ours & \cellcolor{gray!25}57.14 & \cellcolor{gray!25}33.85 & \cellcolor{gray!25}49.32 & \cellcolor{gray!25}44.65 & \cellcolor{gray!25}23.21 & \cellcolor{gray!25}39.86 & \cellcolor{gray!25}47.05 & \cellcolor{gray!25}24.32 & \cellcolor{gray!25}41.57 \\
 & $\text{QD-DETR}^\dag$ \cite{moon2023query} & \underline{59.24} & 33.43 & \underline{50.92} & 42.30 & 21.09 & 38.55 & 46.04 & 26.33 & 42.89 \\
  & \cellcolor{gray!25}QD-DETR+Ours & \cellcolor{gray!25}\textbf{60.66} & \cellcolor{gray!25}\textbf{38.60} & \cellcolor{gray!25}\textbf{52.53} & \cellcolor{gray!25}\textbf{50.23} & \cellcolor{gray!25}\textbf{27.69} & \cellcolor{gray!25}\textbf{44.14} & \cellcolor{gray!25}\textbf{55.25} & \cellcolor{gray!25}\textbf{35.25} & \cellcolor{gray!25}\textbf{48.10} \\
					
\bottomrule
\end{tabular}% End of \resizebox
}
\end{table}

% \cellcolor{gray!25}

\cref{tab:charades} shows the overall performance of our approach on the Charades-CG dataset. We observe that: (1) While the baseline QD-DETR \cite{moon2023query} outperforms the latest state-of-the-art methods (\eg, SSL \cite{li2023exploring} and DeCo\cite{yang2023deco}) in the Test-Trivial spilt, there is still a performance gap in the Novel-Composition and Novel-Word splits, indicating that it has poor compositional generalization capabilities. (2) Our method can significantly improve the compositional generalizability of QD-DETR, elevating 7.93\% and 6.60\% in R1@0.5 and R1@0.7 in Novel-Composition split, respectively, finally outperforming on all three test splits. (3) Our method can be integrated into existing DETR-based models to unlock their compositional generalizability. 
\eg, our approach notably enhances Moment-DETR and improves its performance by 7.66\% and 5.23\% in R1@0.5 in Test-Trivial and Novel-Composition splits, respectively. 

In \cref{tab:anet}, we also achieve competitive results on the ActivityNet-CG dataset across two baselines. In the Novel-Composition split, our method promotes the performance of QD-DETR by 2.65\%, 3.43\%, and 1.43\% in R1@0.5, R1@0.7 and mIoU, respectively. Additionally, by integrating our approach, Moment-DETR's performance experiences a boost of 1.39\%, 0.69\%, and 1.50\% in R1@0.5, R1@0.7 and mIoU metrics, respectively. Notably, VISA \cite{li2022compositional} leverages external knowledge from off-the-shelf object detectors and action recognition models while our method is conducted in an end-to-end manner. Although our method underperforms VISA \cite{li2022compositional} in the Test-Trivial split, it still achieves comparable performance in the Novel-Composition split, which proves that our method has a better capability of compositional generalization.

\renewcommand{\arraystretch}{1.2} % Adjusts the row height
\begin{table}[tb]
\caption{Performance (\%) of state-of-the-art methods on the ActivityNet-CG dataset. The best result is shown in \textbf{bold} and the second best is \underline{underlined}. `WS': weakly-supervised methods. `RL': reinforcement learning methods. `PB': proposal-based methods. `PF': proposal-free methods. $\dag$ indicates the results of our implementation using the officially released code. \textcolor{shade}{$\star$ denotes the results relying on external detector knowledge.}}
\centering
\label{tab:anet}
\scalebox{0.7}{
\begin{tabular}{llccccccccc}
\toprule
\multirow{2}{*}{Setting} & \multirow{2}{*}{Method} & \multicolumn{3}{c}{Test-Trivial} & \multicolumn{3}{c}{Novel-Composition} & \multicolumn{3}{c}{Novel-Word} \\ \cmidrule(lr){3-5} \cmidrule(lr){6-8} \cmidrule(l){9-11}
 & & R1@0.5 & R1@0.7 & mIoU & R1@0.5 & R1@0.7 & mIoU & R1@0.5 & R1@0.7 & mIoU \\  \midrule
WS & WSSL \cite{duan2018weakly} & 11.03 & 4.14 & 15.07 & 2.89 & 0.76 & 7.65 & 3.09 & 1.13 & 7.10 \\  \midrule
RL & TSP-PRL \cite{wu2020tree} & 34.27 & 18.80 & 37.05 & 14.74 & 1.43 & 12.61 & 18.05 & 3.15 & 14.34 \\ \hline
\multirow{3}{*}{PB} & TMN \cite{liu2018temporal} & 16.82 & 7.01 & 17.13 & 8.74 & 4.39 & 10.08 & 9.93 & 5.12 & 11.38 \\
 & 2D-TAN \cite{zhang2020learning} & \textbf{44.50} & \textbf{26.03} & 42.12 & 22.80 & 9.95 & 28.49 & 23.86 & 10.37 & 28.88 \\ 
 \midrule
\multirow{6}{*}{PF} & LGI \cite{mun2020local} & 43.56 & 23.29 & 41.37 & 23.21 & 9.02 & 27.86 & 23.10 & 9.03 & 26.95 \\
 & VLSNet \cite{zhang2020span} & 39.27 & 23.12 & 42.51 & 20.21 & 9.18 & 29.07 & 21.68 & 9.94 & 29.58 \\
 & \textcolor{shade}{$\text{VISA}^\star$} \cite{li2022compositional} & \textcolor{shade}{47.13} & \textcolor{shade}{29.64} & \textcolor{shade}{44.02} & \textcolor{shade}{31.51} & \textcolor{shade}{16.73} & \textcolor{shade}{35.85} & \textcolor{shade}{30.14} & \textcolor{shade}{15.90} & \textcolor{shade}{35.13} \\
 & Deco \cite{yang2023deco} & 43.98 & 24.25 & \underline{43.47} & 27.35 & 11.66 & 31.27 & - & - & - \\ \cmidrule{2-11}
 
 & $\text{Moment-DETR}^\dag$ \cite{lei2021detecting} & 42.73 & 25.31 & 42.19 & 29.29 & 13.71 & 31.63 & 26.84 & \underline{13.34} & 29.95 \\
 & \cellcolor{gray!25}Moment-DETR+Ours & \cellcolor{gray!25}\underline{44.19} & \cellcolor{gray!25}25.81 & \cellcolor{gray!25}\textbf{43.49} & \cellcolor{gray!25}\textbf{30.60} & \cellcolor{gray!25}\textbf{14.40} & \cellcolor{gray!25}\textbf{33.13} & \cellcolor{gray!25}\textbf{29.59} & \cellcolor{gray!25}\textbf{15.10} & \cellcolor{gray!25}\textbf{32.43} \\ 
 & $\text{QD-DETR}^\dag$ \cite{moon2023query} & 41.80 & 20.88 & 41.15 & 26.91 & 10.96 & 31.01 & 27.09 & 11.38 & \underline{31.21} \\
 & \cellcolor{gray!25}QD-DETR+Ours & \cellcolor{gray!25}43.76 & \cellcolor{gray!25}\underline{25.98} & \cellcolor{gray!25}42.86 & \cellcolor{gray!25}\underline{29.56} & \cellcolor{gray!25}\underline{14.37} & \cellcolor{gray!25}\underline{32.44} & \cellcolor{gray!25}\underline{27.60} & \cellcolor{gray!25}13.11 & \cellcolor{gray!25}30.98 \\
\bottomrule

\end{tabular}
}
\end{table}

\begin{table}[tb]
\caption{Ablation studies for coarse-grained ranking on Charades-CG dataset. * means that we replace the saliency loss in $\mathcal{L}_{base}$ with our $\mathcal{L}_{intra}$ and yield better performance.}
\label{tab:charades-coarse}
\centering
\scalebox{0.7}{ % Adjust scale as needed
\begin{tabular}{ccc|ccc|ccc}
\toprule
\multirow{2}{*}{$\mathcal{L}_{base}$} & \multirow{2}{*}{$\mathcal{L}_{intra}$} & \multirow{2}{*}{$\mathcal{L}_{inter}$} & \multicolumn{3}{c|}{Test-Trivial} & \multicolumn{3}{c}{Novel-Composition}\\ 
\cmidrule(lr){4-6} \cmidrule(lr){7-9}
& & & R1@0.5 & R1@0.7 & mIoU & R1@0.5 & R1@0.7 & mIoU\\  
 \midrule
\checkmark &  &  & 59.24 & 33.43 & 50.92 & 42.30 & 21.09 & 38.55 \\ 
\checkmark & \checkmark* &  & 60.24 & 35.89 & 51.73 & 44.02 & 22.84 & 39.23 \\
\checkmark & & \checkmark & 60.17 & 37.11 & 51.96 & \textbf{46.69} & 24.87 & 41.74\\
\checkmark & \checkmark* & \checkmark & \textbf{61.98} & \textbf{37.56} & \textbf{53.38} & 46.25 & \textbf{24.93} & \textbf{41.88} \\
\bottomrule
\end{tabular}%
}
\end{table}

\subsection{Ablation Studies}
We further provide ablation studies to validate the effectiveness of the proposed method, including various constraints in the coarse-to-fine saliency ranking loss, diverse hierarchical negative queries, and several hyperparameter settings. We use QD-DETR as the baseline to explore the insights.

\noindent \textbf{Coarse-Grained Saliency Ranking.} 
We report the contributions of each constraint within the coarse-grained saliency ranking loss in \cref{tab:charades-coarse}. Note that the proposed intra-ranking loss shares the same objective with the saliency loss in $\mathcal{L}_{base}$. Therefore, by replacing it with $\mathcal{L}_{intra}$,  our loss yields improvements across all metrics on two test sets, with R1@0.7 increasing by 2.46\% on Test-trivial and 1.75\% on Novel-Composition. We also notice that the boosting effect of $\mathcal{L}_{inter}$ is more significant compared to $\mathcal{L}_{intra}$. By combining both with $\mathcal{L}_{base}$, the overall performance can be further increased, resulting in absolute gains of 2.46\% and 3.33\% for mIoU on Test-Trivial and Novel-Composition, respectively. The results show the $\mathcal{L}_{cr}$ enhances the discriminability between video clips inside and outside the ground truth interval, as well as between positive and negative query responses, simultaneously improving the compositional generalizability.

\noindent \textbf{Fine-Grained Saliency Ranking.} 
We present the effects of each constraint within the fine-grained saliency ranking loss in \cref{tab:charades-fine}, and its synergy with coarse-grained ranking loss. We can observe that: (1) As fine-grained ranking constraints are gradually added, there is a general trend of improvement across all three metrics, among which $\mathcal{L}^1_{fr}$ plays a leading role, significantly improving R1@0.5 by 4.13\%. (2) Without complete constraints, the introduction of $\mathcal{L}^3_{fr}$ leads to a slight performance degradation. This trend remains consistent both before and after combining with $\mathcal{L}_{cr}$. Interestingly, when all constraints are integrated into the baseline, $\mathcal{L}^3_{fr}$ further improves the performance by 3.17\% in R1@0.7, which suggests that it only realizes full potential when forming a complete hierarchy of constraints.

\noindent \textbf{Comparison of Hierarchical Negative Queries.} We compare the effects of the random sample-based and LLM-based hard negative queries in \cref{tab:negatives}. From the Charades-CG results, we observe that LLM-based negative queries significantly outperform random sampling, with improvements of 2.82\% on R1@0.5 and 1.64\% on mIoU. 

Among all the evaluated LLMs, the negative samples generated by GPT-3.5 Turbo perform better.
We also note that the improvement using the LLM-based approach over random sampling is less significant in ActivityNet-CG. A possible reason is that the query sentences are longer while the replacement ratio is lower, so the retained context is still sufficient for localization. However, the LLM-based approach consistently outperforms random sampling in R1@0.7, indicating its potential for precise grounding.

\begin{table}[tb]
    \begin{minipage}[t]{0.50\linewidth} % Adjusts alignment to top
    \centering
    \caption{Ablation studies for fine-grained saliency ranking constraints in the Novel-Composition split of Charades-CG. $\mathcal{L}^1_{fr}$ to $\mathcal{L}^4_{fr}$ are four constraints in order in $\mathcal{L}_{fr}$.}
    \scalebox{0.55}{
    \begin{tabular}{c@{\hspace{10pt}}c@{\hspace{10pt}}c@{\hspace{10pt}}c@{\hspace{10pt}}c@{\hspace{10pt}}c@{\hspace{10pt}}|c@{\hspace{10pt}}c@{\hspace{12pt}}c}
    \toprule
    $\mathcal{L}_{base}$ & $\mathcal{L}^1_{fr}$ & $\mathcal{L}^2_{fr}$ & $\mathcal{L}^3_{fr}$ & $\mathcal{L}^4_{fr}$ & $\mathcal{L}_{cr}$ & R1@0.5 & R1@0.7 & mIoU \\  \midrule
    \checkmark &  & &  & & & 42.30 & 21.09 & 38.55 \\
    \checkmark & \checkmark & &  & & & 46.43 & 23.39 & 41.65 \\
    \checkmark & \checkmark & \checkmark &  & & & 46.72 & 24.35 & 42.29 \\
    \checkmark & \checkmark & \checkmark & \checkmark  & & & 46.40 & 23.30 & 41.88 \\
    \checkmark & \checkmark  & \checkmark & \checkmark  & \checkmark & & 46.98 & 24.00 & 41.94 \\ \midrule
    \checkmark & \checkmark & \checkmark &  & & \checkmark & 48.40 & 24.55 & 43.05 \\
    \checkmark & \checkmark & \checkmark & \checkmark &  & \checkmark & 47.41 & 25.68 & 41.52 \\
    \checkmark & \checkmark & \checkmark & & \checkmark & \checkmark & 49.51 & 24.52 & 43.27 \\
    \checkmark & \checkmark & \checkmark & \checkmark & \checkmark & \checkmark & \textbf{50.23} & \textbf{27.69} & \textbf{44.14} \\
    \bottomrule
    \end{tabular}
    }
    \label{tab:charades-fine}
    \end{minipage}% End of the left-hand side minipage
    \hfill % Adds space between the minipages
    \begin{minipage}[t]{0.45\linewidth} % Adjusts alignment to top and starts the right-hand side minipage
    \centering
    \caption{Comparison of different large language models for hard negative construction in the Novel-Composition test split.}
    \scalebox{0.6}{
    \begin{tabular}{c|l|ccc}
    \toprule
    Dataset & Hard Negatives & R1@0.5 & R1@0.7 & mIoU \\  \midrule
    \multirow{4}{*}{Charades-CG} & random sample  & 47.41 & 25.33 & 42.50 \\ 
    & Llama 3~\cite{llama3}  & 48.75 & 25.22 & 42.89 \\
    & Gemini-1.5 Flash~\cite{gemini} & 48.69 & 25.60 & 43.54 \\
     & GPT-3.5 Turbo~\cite{brown2020language} & \textbf{50.23} & \textbf{27.69} & \textbf{44.14} \\
    \midrule
    \multirow{4}{*}{ActivityNet-CG} & random sample & 29.59 & 12.89 & 32.06 \\
    & Llama 3~\cite{llama3} & 29.24 & 13.56 & 31.82 \\
    & Gemini-1.5 Flash~\cite{gemini} & \textbf{29.72} & 13.24 & 31.96 \\
    & GPT-3.5 Turbo~\cite{brown2020language}   & 29.56 & \textbf{14.37} & \textbf{32.44} \\
    \bottomrule
    \end{tabular}
    }
    \label{tab:negatives}
    \end{minipage} % End of the right-hand side minipage
\end{table}

\begin{table}[tb]
    \centering
    \begin{minipage}[t]{0.48\linewidth} % Adjust the width for better alignment
        \centering
        \caption{Contribution of including \emph{prepositions} and \emph{adverbs} into hard negatives on Charades-CG Novel Composition split.}
        \scalebox{0.65}{ % Adjust scale to fit the table within the page
            \begin{tabular}{l|ccc}
                \toprule
                Method & R1@0.5 & R1@0.7 & mIoU \\  
                \midrule
                MD+Ours w/o \emph{prep} \& \emph{adv} & 43.03 & 22.08 & 38.79 \\
                MD+Ours & 44.65 & 23.21 & 39.86 \\ \hline
                QD+Ours w/o \emph{prep} \& \emph{adv} & 48.87 & 25.28 & 43.30 \\
                QD+Ours & \textbf{50.23} & \textbf{27.69} & \textbf{44.14} \\
                \bottomrule
            \end{tabular}%
        }
        \label{tab:prep-adv-ablation}
    \end{minipage}%
    \hspace{0.02\linewidth} % Reduces space between the minipages
    \begin{minipage}[t]{0.48\linewidth} % Adjust the width for better alignment
        \centering
        \caption{Performance (mIoU) of our method on different composition types of Charades-CG Novel Composition split.}
        \scalebox{0.65}{ % Adjust scale to fit the table within the page
            \begin{tabular}{l|ccccc}
                \toprule
                Method & verb-noun & adj-noun & noun-noun & verb-adv & prep-noun \\  
                \midrule
                MD & 36.01 & 28.79 & 41.95 & 34.45 & 35.38\\
                MD+Ours & 40.29 & 37.54 & 41.56 & 37.69 & 38.48 \\ \hline
                QD & 37.51 & 41.14 & 44.70 & 34.65 & 33.72 \\
                QD+Ours & 43.25 & 45.34 & 45.53 & 41.05 & 38.71 \\
                \bottomrule
            \end{tabular}%
        }
        \label{tab:different-compositions}
    \end{minipage}
\end{table}

\noindent \textbf{Ablation of Different Composition Types.}
We conduct an ablation study by excluding \emph{prepositions} and \emph{adverbs} for hard negative construction. \cref{tab:prep-adv-ablation} shows that considering \emph{prepositions} and \emph{adverbs} effectively improves the model’s perception of non-dominant primitives. \cref{tab:different-compositions} further shows that our method consistently improves the compositional generalizability of existing DETR-based methods across different composition types, validating the rationality of the proposed hard negative construction method.

\begin{figure}[tb]
  \centering
  \begin{subfigure}{0.49\linewidth}
    \includegraphics[width=\textwidth]{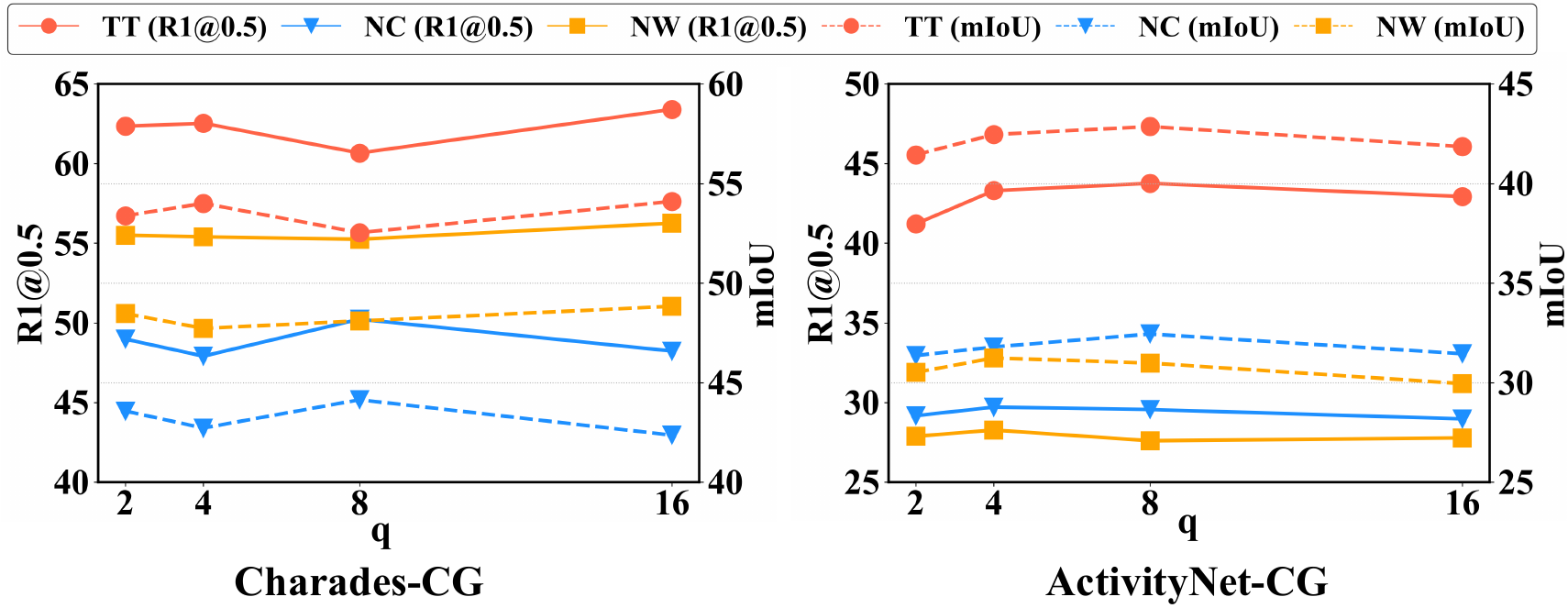}
    \caption{The effect of different $q$ in top-$k$ selection for coarse-grained ranking loss.}
    \label{fig:topk-q}
  \end{subfigure}
  \hspace{5pt}
  \begin{subfigure}{0.47\linewidth}
    \includegraphics[width=\textwidth]{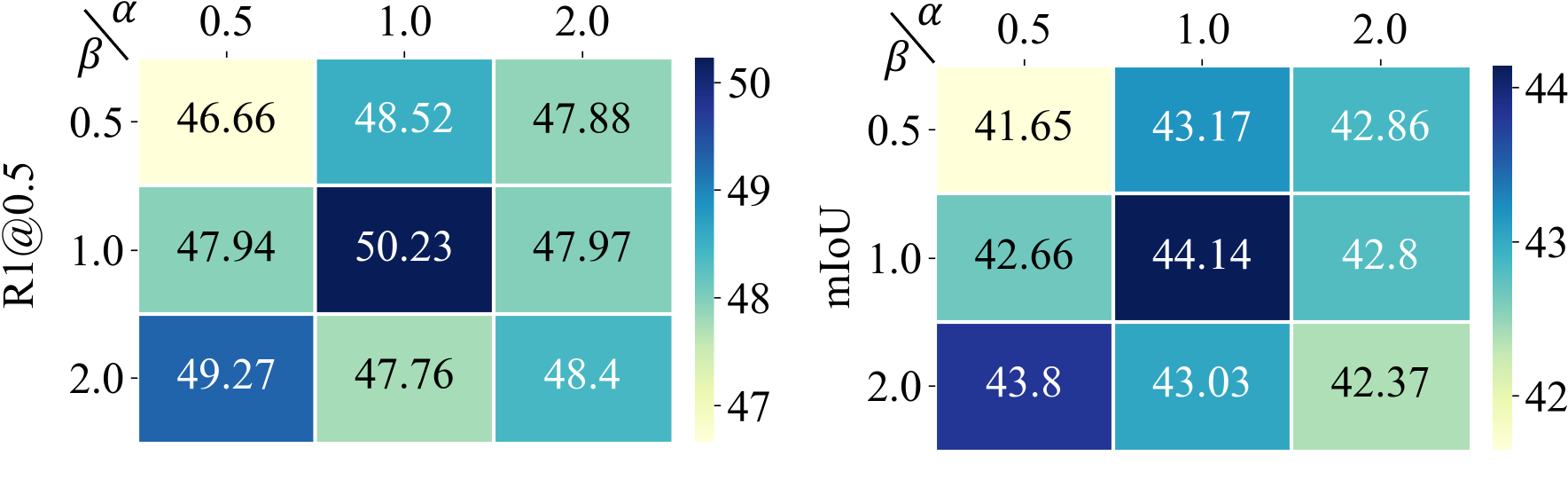}
    \caption{Comparison of the trade-off loss weights in the Novel-Composition split of the Charades-CG Dataset.}
    \label{fig:loss-weight}
  \end{subfigure}
  \caption{Hyperparameter Evaluation.}
  \label{fig:fig5}
\end{figure}

\noindent \textbf{Hyperparameter Evaluation.} 
In \cref{fig:topk-q}, we explore the effect of different $q$ on the top-$k$ selection in \cref{eq:topk}. For Charades-CG, with the increase of $q$, the metrics including R1@0.5 and mIoU on Test-Trivial and Novel-Composition show an opposite trend, indicating a competitive balance between the two splits. For ActivityNet-CG, the trends for Test-Trivial and Novel-Composition are consistent, while larger $q$ leads to a decrease in performance on all three splits. When $q=8$, the model achieves the best Novel-Composition results on both datasets. The heatmaps in \cref{fig:loss-weight} illustrate the effects of different loss weights $\alpha$ and $\beta$ in \cref{eq:overall} in the Novel-Composition split. We empirically find that the optimal performance in R1@0.5 and mIoU is achieved when both $\alpha$ and $\beta$ are set to 1.0. Therefore, we use this setting by default in our experiments.

\begin{figure}[tb]
  \centering
  \begin{subfigure}{0.48\linewidth}
    \includegraphics[width=\textwidth]{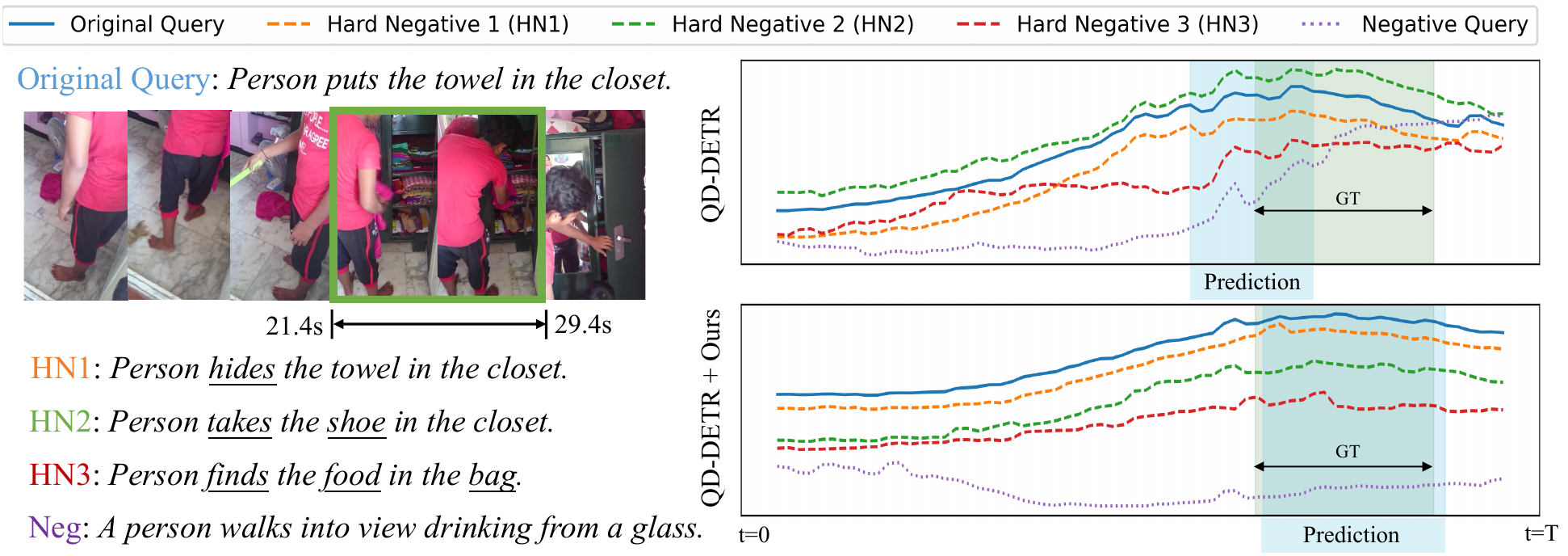}
    \caption{} % T8B7B.mp4
    \label{fig:T8B7B}
  \end{subfigure}
  \begin{subfigure}{0.48\linewidth}
    \includegraphics[width=\textwidth]{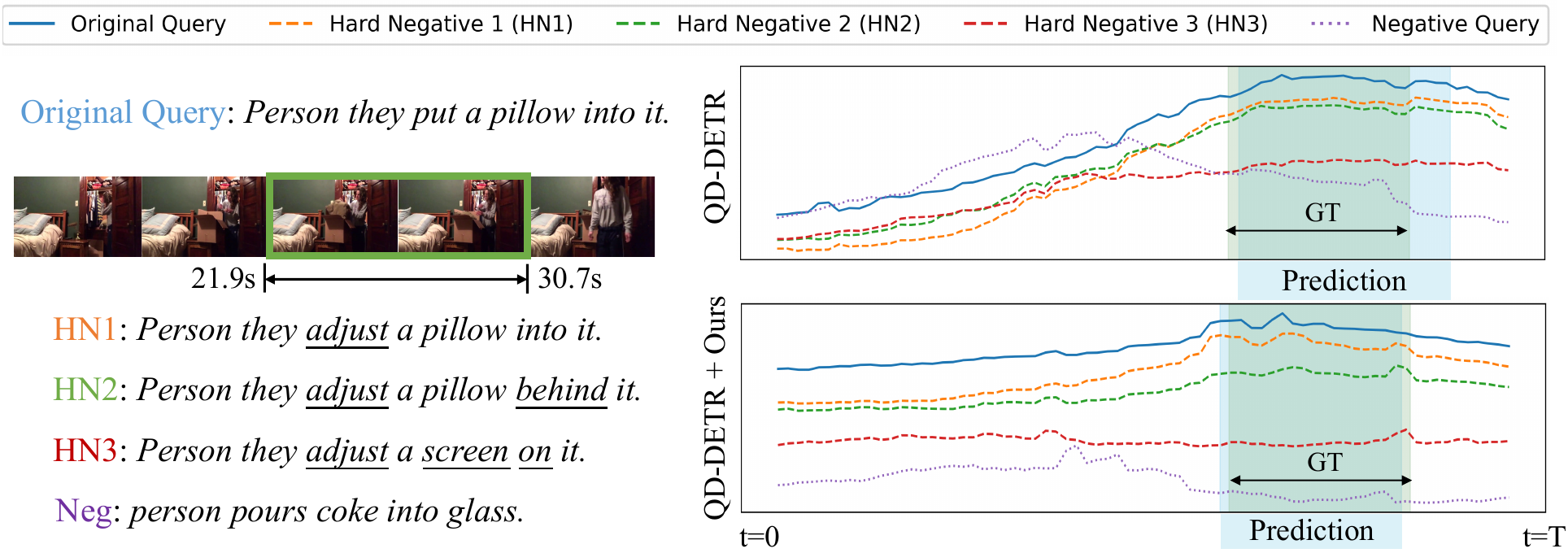}
    \caption{}  % BID6Z.mp4
    \label{fig:BID6Z}
  \end{subfigure}
  \caption{Visualization of saliency scores given different query sentences. (a) and (b) are test samples from Charades-CG.}
  \label{fig:cc-saliency}
\end{figure}

\begin{figure}[tb]
  \centering
  \begin{subfigure}{\linewidth}
    \includegraphics[width=\textwidth]{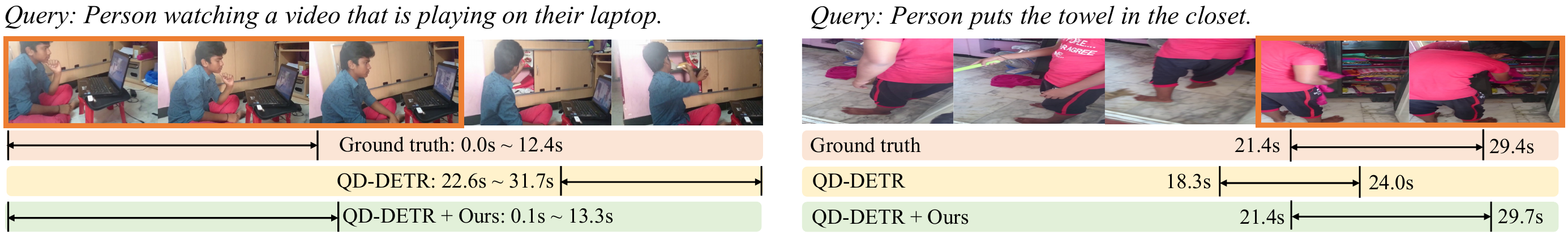}
    \caption{Two Samples from the Test-Trivial split}
    \label{fig:cc-tt}
  \end{subfigure}
  \begin{subfigure}{\linewidth}
    \includegraphics[width=\textwidth]{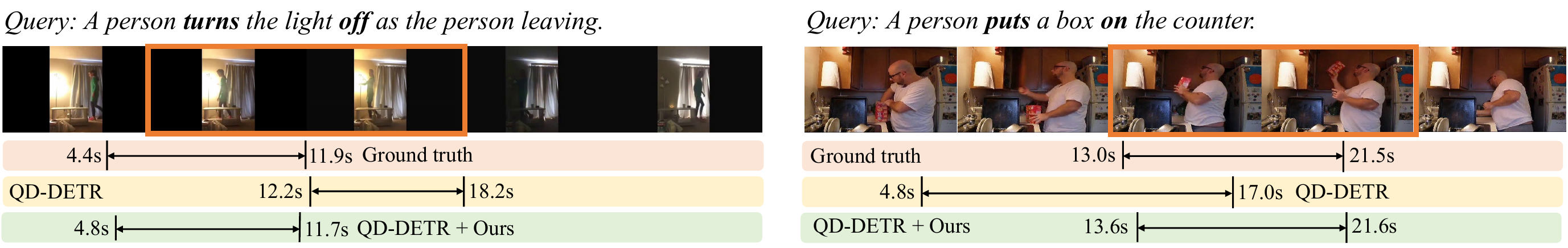}
    \caption{Two Samples from the Novel-Composition split}
    \label{fig:cc-nc}
  \end{subfigure}
  \begin{subfigure}{\linewidth}
    \includegraphics[width=\textwidth]{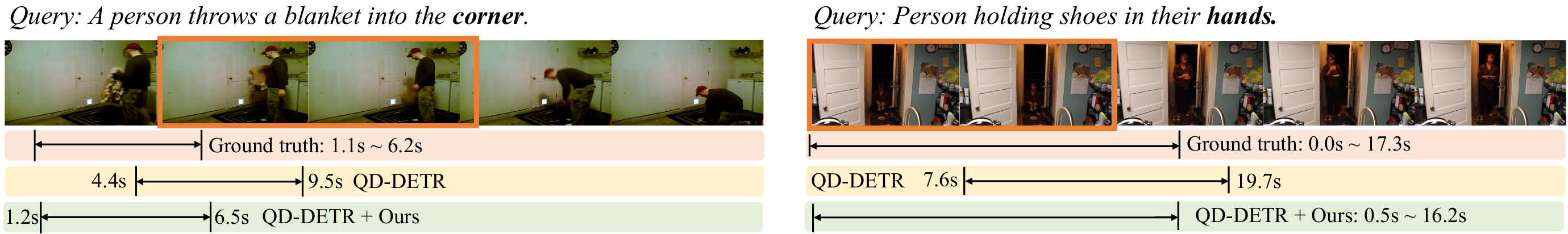}
    \caption{Two Samples from the Novel-Word split}
    \label{fig:cc-nw}
  \end{subfigure}
  \caption{Qualitative comparison in different test splits of Charades-CG.}
  \label{fig:visual}
\end{figure}

\subsection{Qualitative Analysis}

We visualize the saliency scores of several cases in Charades-CG in \cref{fig:cc-saliency} and observe that the existing work struggles with hard negative queries, showing irrational saliency responses. For instance, in \cref{fig:T8B7B}, the hard negative query “\textit{Person \underline{takes} the \underline{shoe} in the closet.}” is even more salient than the positive query “\textit{Person puts the towel in the closet}”, leading to imprecise moment localization. In contrast, our approach consistently improves the model’s ability to distinguish different words between positive and hard negative queries and yield hierarchical responses, thereby achieving better moment localization and compositional generalization. Additionally, our method can also catch the nuanced variation of adverbs and prepositions, such as different prepositions “into” and “behind” in \cref{fig:BID6Z}. This indicates that ours are more sensitive to the semantic changes of different non-dominant primitives. 

\cref{fig:visual} illustrates several qualitative examples in three splits of Charades-CG. In the Test-Trivial split, although the queries don’t contain unseen compositions and words, our method demonstrates more precise alignment than the baseline. When encountering the Novel-composition "\textit{\textbf{turns} the light \textbf{off}}" and “\textit{\textbf{put} a box \textbf{on}}” and Novel-Word “\textit{\textbf{corner}”} and \textit{“\textbf{hands}}” in the queries, our method can still generalise well to them. The presented results indicate that our method effectively guides DETR-based models in utilizing hierarchical negative samples to enhance the generalizability of unseen compositions and unseen words. 

\section{Conclusion}
In this paper, we propose SHINE, a Saliency-aware HIerarchical NEgative ranking method for compositional temporal grounding. We first utilize an LLM to produce semantically plausible yet distinct hierarchical hard negatives from the original query. Furthermore, we introduce a coarse-to-fine saliency ranking strategy that establishes a multi-granularity semantic relationship between video and hard negatives. Extensive experiments demonstrate that SHINE substantially enhances the compositional generalization capabilities of current DETR-based temporal grounding models.

\section*{Acknowledgements}
This work was partially supported by Veritone and Adobe and has utilised Queen Mary’s Apocrita HPC facility from QMUL Research-IT. This work was also partially supported by the Office of Naval Research (ONR) grant (N00014-23-1-2417 \& N00014-23-1-2046), National Science Foundation (NSF) CAREER under award 2028626, and NSF SaTC Award 1949694. Any opinions, findings, and conclusions or recommendations expressed in this material are those of the authors and do not necessarily reflect the views of ONR or NSF.

\bibliographystyle{splncs04}
\bibliography{main}

\clearpage

\appendix
\section{Appendix}
\label{appendix}
\subsection{Supplementary Experiments}
In this section, we provide more supplementary experimental results to explore: (1) different masking ratios for hard negative construction; (2) different ranking margins in coarse- and fine-grained saliency ranking loss; (3) different variants of fine-grained saliency ranking loss; (4) more details of different $\mathcal{L}_{base}$ in Moment-DETR and QD-DETR; and (5) performance on a classic temporal grounding benchmark.

\subsubsection{Different Masking Ratios for Hard Negative Construction}

In this subsection, we explore the performance when applying different masking ratios to generate hard negative queries. We report the results of QD-DETR with our methods on ActivityNet-CG in \cref{tab:mask}. From the results we can observe that: (1) Larger masking ratios can yield better performance in both Test-Trivial and Novel-Word splits. (2) The model's performance in the Novel-Composition split is more sensitive to different masking ratios. (3) `QD-DETR + Ours' achieves the best overall performance on the three test splits with the masking ratios of 10\%, 30\%, and 50\%, which we keep as our default settings in all experiments.

\begin{table}[b]
\caption{Effect of different masking ratios on the ActivityNet-CG dataset. $r_1$ to $r_3$ denote the progressive masking ratio for hard negative construction.}
\label{tab:mask}
\centering
\scalebox{0.8}{ % Adjust scale as needed
\begin{tabular}{c@{\hspace{10pt}}c@{\hspace{10pt}}c@{\hspace{10pt}}|ccc|ccc|ccc}
\toprule
\multirow{2}{*}{$r_1$} & \multirow{2}{*}{$r_2$} & \multirow{2}{*}{$r_3$} & \multicolumn{3}{c|}{Test-Trivial} & \multicolumn{3}{c|}{Novel-Composition} & \multicolumn{3}{c}{Novel-Word}\\ 
\cmidrule(lr){4-6} \cmidrule(lr){7-9} \cmidrule(lr){10-12}
& & & R1@0.5 & R1@0.7 & mIoU & R1@0.5 & R1@0.7 & mIoU & R1@0.5 & R1@0.7 & mIoU\\  
 \midrule
0.10 & 0.30 & 0.50 & 43.76 & \textbf{25.98} & 42.86 & \textbf{29.56} & \textbf{14.37} & \textbf{32.44} & 27.60 & \textbf{13.11} & 30.98 \\ 
% 0.20 & 0.40 & 0.60 & - & - & - & - & - & - & - & - & - \\ 
0.25 & 0.50 & 0.75 & 43.49 & 25.63 & 42.82 & 28.42 & 12.94 & 31.20 & 27.40 & 12.73 & 30.66  \\
0.30 & 0.60 & 0.90 & \textbf{44.04} & 25.95 & \textbf{43.13} & 29.41 & 13.51 & 32.02 & \textbf{28.11} & 12.78 & \textbf{31.30} \\
\bottomrule
\end{tabular}%
}
\end{table}

\subsubsection{Different Ranking Margins in Coarse- and Fine-grained Saliency Ranking loss}
% In the submission, we adopt the cross-entropy between two saliency scores derived from adjacent hard negatives to construct fine-grained ranking loss, as illustrated in \cref{eq:fine}.
We also explore the effectiveness of different margins in the coarse- and fine-grained saliency ranking loss. For the coarse-grained ranking margin, \cref{tab:charades-coarse-margin} shows that increasing intra-margin $h_1$ and inter-margin $h_2$ enhances the model performance, peaking when $h_1=1.0$ and $h_2=2.0$. Regarding the fine-grained ranking margins $m_0$ to $m_3$, \cref{tab:cc-relative} shows the highest results achieved by setting all margins in $\mathcal{L}_{fr}$ to 0.25. By choosing the appropriate margins, our method leads the model to learn nuances between different negative queries through $\mathcal{L}_{cr}$ and $\mathcal{L}_{fr}$, which enables the model to alleviate irrational saliency responses and improves the compositional generalizability. We then provide more comparative results under the setting of different fine-grained ranking margins. It can be seen that: (1) As the four margin values increase, the overall performance of the Novel-Composition split initially rises and then falls, with the best result achieved at 0.25. (2) The R1@0.7 metric is more sensitive to changes in margin values, indicating that appropriate margin values are beneficial for precise localization.

\begin{table}[tb]
    \caption{Ablation of different margins in the Novel-Composition split of Charades-CG.}
    % \vspace{-10pt}
    \centering
    \begin{subtable}{.35\linewidth}
      \centering
        \caption{Intra and inter margins in $\mathcal{L}_{cr}$}
        \label{tab:charades-coarse-margin}
        \scalebox{0.75}{
            \begin{tabular}{c@{\hspace{10pt}}c@{\hspace{5pt}}|cc@{\hspace{5pt}}c}
                \toprule
                $h_1$ & $h_2$ & R1@0.5 & R1@0.7 & mIoU \\  
                \midrule
                0.2 & 1.0 & 47.91 & 24.35 & 42.76 \\
                0.5 & 1.0 & 47.04 & 26.03 & 42.29 \\
                1.0 & 1.0 & 48.63 & 24.35 & 42.98 \\
                1.0 & 2.0 & \textbf{50.23} & \textbf{27.69} & \textbf{44.14} \\
                \bottomrule
            \end{tabular}%
        }
    \end{subtable}
    \begin{subtable}{.4\linewidth}
      \centering
        \caption{Relative ranking margins in $\mathcal{L}_{fr}$}
        \label{tab:cc-relative}
        \scalebox{0.75}{
            \begin{tabular}{c@{\hspace{10pt}}c@{\hspace{10pt}}c@{\hspace{10pt}}c@{\hspace{5pt}}|cc@{\hspace{5pt}}c}
                \toprule
                $m_0$ & $m_1$ & $m_2$ &$m_3$ & R1@0.5 & R1@0.7 & mIoU \\  
                \midrule
                0.10 & 0.10 & 0.10 & 0.10 & 49.27 & 25.54 & 43.93 \\
                0.20 & 0.20 & 0.20 & 0.20 & 48.63 & 27.31 & 43.15 \\
                0.25 & 0.25 & 0.25 & 0.25 & \textbf{50.23} & \textbf{27.69} & \textbf{44.14} \\
                0.50 & 0.50 & 0.50 & 0.50 & 47.18 & 25.57 & 41.83 \\
                \midrule
                0.05 & 0.05 & 0.15 & 0.25 & 49.27 & 24.46 & 43.48 \\
                0.05 & 0.10 & 0.15 & 0.20 & 49.07 & \textbf{25.86} & 43.59 \\
                0.10 & 0.30 & 0.50 & 0.70 & 48.02 & 25.10 & 42.50 \\
                0.25 & 0.50 & 0.75 & 1.0 & \textbf{50.00} & 24.87 & \textbf{43.78} \\
                \bottomrule
            \end{tabular}%
        }
    \end{subtable}
    % \vspace{-15pt}
\end{table}
% \vspace{-15pt}

\subsubsection{Different Variants of Fine-grained Saliency Ranking Loss}

Moreover, we also explore an absolute ranking regime with incremental margins from $m_0$ to $m_3$, denoted as \cref{eq:fine_var}, where we fix the observations in the second through fourth constraints to be $d(S_p, S^1_{hn})$. From \cref{tab:cc-absolute} we notice that this variant can also achieve considerable improvements, while our vanilla version still achieves optimal results, which we attribute to its finer constraints on adjacent hard negative queries to capture nuances. Therefore, we adopt the fine-grained ranking loss of relative distance as the default setting in all experiments.

% \begin{equation}
% \begin{aligned}
% \mathcal{L}_{fr} = & \max(0, m_0 + d(Y, S_p) - d(S_p, S^1_{hn})) \\
% & + \max(0, m_1 + d(S_p, S^1_{hn}) - d(S_p, S^2_{hn})) \\
% & + \max(0, m_2 + d(S_p, S^2_{hn}) - d(S_p, S^3_{hn})) \\
% & + \max(0, m_3 + d(S_p, S^3_{hn}) - d(S_p, S_{n})),
% \end{aligned}
% \label{eq:fine}
% \end{equation}

\begin{equation}
\begin{aligned}
\mathcal{L}_{fr}' = & \max(0, m_0 + d(Y, S_p) - d(S_p, S^1_{hn})) \\
& + \max(0, m_1 + d(S_p, S^1_{hn}) - d(S_p, S^2_{hn})) \\
& + \max(0, m_2 + d(S_p, S^1_{hn}) - d(S_p, S^3_{hn})) \\
& + \max(0, m_3 + d(S_p, S^1_{hn}) - d(S_p, S_{n})),
\end{aligned}
\label{eq:fine_var}
\end{equation}

\begin{table}[tb]
    \caption{Absolute ranking margins in $\mathcal{L}_{fr}'$ in the Novel-Composition split of the Charades-CG Dataset.}
    % \vspace{-10pt}
    % \centering
    % \begin{subtable}{.4\linewidth}
    %   \centering
    %     \caption{Relative ranking margins in $\mathcal{L}_{fr}$}
    %     \label{tab:cc-relative}
    %     \scalebox{0.75}{
    %         \begin{tabular}{c@{\hspace{10pt}}c@{\hspace{10pt}}c@{\hspace{10pt}}c@{\hspace{5pt}}|cc@{\hspace{5pt}}c}
    %             \toprule
    %             $m_0$ & $m_1$ & $m_2$ &$m_3$ & R1@0.5 & R1@0.7 & mIoU \\  
    %             \midrule
    %             0.10 & 0.10 & 0.10 & 0.10 & 49.27 & 25.54 & 43.93 \\
    %             0.20 & 0.20 & 0.20 & 0.20 & 48.63 & 27.31 & 43.15 \\
    %             0.25 & 0.25 & 0.25 & 0.25 & \textbf{50.23} & \textbf{27.69} & \textbf{44.14} \\
    %             0.50 & 0.50 & 0.50 & 0.50 & 47.18 & 25.57 & 41.83 \\
    %             \midrule
    %             0.05 & 0.05 & 0.15 & 0.25 & 49.27 & 24.46 & 43.48 \\
    %             0.05 & 0.10 & 0.15 & 0.20 & 49.07 & \textbf{25.86} & 43.59 \\
    %             0.10 & 0.30 & 0.50 & 0.70 & 48.02 & 25.10 & 42.50 \\
    %             0.25 & 0.50 & 0.75 & 1.0 & \textbf{50.00} & 24.87 & \textbf{43.78} \\
    %             \bottomrule
    %         \end{tabular}%
    %     }
    % \end{subtable}
    % \hspace{20pt}
    % \begin{subtable}{.4\linewidth}
      \centering
        % \caption{Absolute ranking margins in $\mathcal{L}_{fr}'$}
        \label{tab:cc-absolute}
        \scalebox{0.8}{
            \begin{tabular}{c@{\hspace{10pt}}c@{\hspace{10pt}}c@{\hspace{10pt}}c@{\hspace{5pt}}|cc@{\hspace{5pt}}c}
                \toprule
                $m_0$ & $m_1$ & $m_2$ &$m_3$ & R1@0.5 & R1@0.7 & mIoU \\  
                \midrule
                0.05 & 0.10 & 0.15 & 0.20 & 47.62 & 25.19 & 42.49 \\
                0.10 & 0.15 & 0.30 & 0.45 & 48.40 & \textbf{26.67} & 43.15 \\
                0.10 & 0.20 & 0.30 & 0.40 & 47.73 & 25.51 & 43.21 \\
                0.10 & 0.30 & 0.50 & 0.70 & \textbf{49.45} & 25.22 & \textbf{44.02} \\
                0.25 & 0.50 & 0.75 & 1.0 & 47.18 & 24.23 & 42.07 \\
                \bottomrule
            \end{tabular}%
        }
    % \end{subtable}
    % \vspace{-15pt}
\end{table}

\subsubsection{Details of Different $\mathcal{L}_{base}$ in Moment-DETR and QD-DETR}

In our experiments, we retain all the original loss functions in our two baseline models, Moment-DETR~\cite{lei2021detecting} and QD-DETR~\cite{moon2023query}, with the exception of the saliency loss. 
We replaced the original saliency loss with the proposed $\mathcal{L}_{intra}$ due to its better performance. Specifically, for moment retrieval, the $\mathcal{L}_{base}$ in Moment-DETR includes the span loss (L1 Loss + GIoU loss), as defined in \cref{eq:loss_span} and combined with the classification loss (Negative Log-Likelihood loss). In addition, we borrow the negative pair loss (\cref{eq:loss_neg}) from QD-DETR to compute the negative saliency loss for an easy negative query. 

\begin{equation}
\begin{aligned}
\mathcal{L}_{span} = & \lambda_{L1}|| m - \hat{m} || + \lambda_{GIoU} \mathcal{L}_{GIoU}(m - \hat{m}) 
\end{aligned}
\label{eq:loss_span}
\end{equation}

\begin{equation}
\begin{aligned}
\mathcal{L}_{neg} = & - \log(1 - S_{neg})
\end{aligned}
\label{eq:loss_neg}
\end{equation}
Therefore, the $\mathcal{L}_{base}$ of moment-DETR can be formulated as follow:

\begin{equation}
\begin{aligned}
\mathcal{L}_{base_\text{moment-DETR}} = & \lambda_{neg}\mathcal{L}_{neg} + \sum_{i=1}^{N} \left[ -\lambda_{\text{cls}} \log \hat{p}_{\hat{m}} (c_i) + {L}_{span} \right]
\end{aligned}
\label{eq:loss_moment}
\end{equation}
where $m$ and $\hat{m}$ denotes the ground truth and predicted moments, respectively, $N$ denotes the number of moment queries, and $\lambda_{neg}, \lambda_{cls}, \lambda_{L1}$ and $\lambda_{GIoU} \in \mathbb{R}$ are weights balancing the loss. In terms of QD-DETR~\cite{moon2023query}, an extra contrastive ranking loss is included to learn the precisely segmented saliency levels, denoted as:
\begin{equation}
\begin{aligned}
\mathcal{L}_{cont} = & - \sum_{r=1}^R \log \left( \frac{ \sum_{x \in X_{\text{pos}}^{\text{r}}} \exp \left( \frac{S(x)}{\tau} \right) }{ \sum_{x \in (X_{\text{pos}}^r \cup X_{\text{neg}}^{\text{r}})} \exp \left( \frac{S(x)}{\tau} \right) } \right)
\end{aligned}
\label{eq:loss_cont}
\end{equation}
where $R$ denotes the maximum rank value, $X_{r}^{pos}$ and $X_{r}^{neg}$ denotes the positive/negative set in the $r^{th}$ iteration, and $S$ denotes the saliency score.
The overall $\mathcal{L}_{base}$ of moment-DETR can be formulated as:

\begin{equation}
\begin{aligned}
\mathcal{L}_{base_\text{QD-DETR}} = & \lambda_{neg}\mathcal{L}_{neg} + \lambda_{cont}\mathcal{L}_{cont} + \sum_{i=1}^{N} \left[ -\lambda_{\text{cls}} \log \hat{p}_{\hat{m}} (c_i) + {L}_{span} \right]
\end{aligned}
\label{eq:loss_qd}
\end{equation}
where $\lambda_{cont} \in \mathbb{R}$ is the contrastive loss weights. For more details, please refer to their original papers~\cite{lei2021detecting, moon2023query}.

\subsubsection{Performance on Charades-STA}
To further validate the applicability and compatibility of our method, we conducted additional experiments on a widely used temporal grounding benchmark, \ie, Charades-STA \cite{gao2017tall}. The training set contains 12,404 queries with 5,336 videos and the test set consists of 3,720 queries with 1,334 videos, which is used for evaluating IID (Independent and Identically Distributed) generalization capability.

% ActivityNet Captions \cite{krishna2017dense} contains 20,000 videos and 100,000 segment-caption pairs. Following \cite{mun2020local, yan2023unloc}, we use train split for training, val\_1 for validation and val\_2 for testing. 

From \cref{tab:c-sta} we can observe that: (1) Our method can consistently improve the performance of two baselines, \ie, Moment-DETR and QD-DETR, with 3.62\% and 2.5\% absolute gain in R1@0.5, respectively. (2) QD-DETR with our method achieves the new state-of-the-art results in both R1@0.5 and R1@0.7, while achieving comparable results to HISA \cite{xu2022hisa} in mIoU. Notably, our reproduced results for QD-DETR$^\dag$ are significantly higher than those reported in their paper. Since they did not provide detailed training settings, we followed Moment-DETR's hyperparameters in addition to the learning rate, which is consistent with that in our submission.

\renewcommand{\arraystretch}{1.2} % Adjusts the row height
\begin{table}[tb]
\caption{Performance (\%) of state-of-the-art methods on the Charades-STA dataset. Our results are shown in \textbf{bold}. `RL': reinforcement learning methods. `PB': proposal-based methods. `PF': proposal-free methods. $\dag$ indicates the results of our re-implementation using the officially released code. - indicates results are not available.}
\centering
\label{tab:c-sta}
\scalebox{0.7}{
\begin{tabular}{l@{\hspace{5pt}}l@{\hspace{10pt}}c@{\hspace{10pt}}c@{\hspace{5pt}}c@{\hspace{5pt}}c@{\hspace{5pt}}}
\toprule
Setting & Method & Feature & R1@0.5 & R1@0.7 & mIoU\\ 
% \cmidrule(lr){3-5} \cmidrule(lr){6-8} \cmidrule(l){9-11}
\midrule
RL & TSP-PRL \cite{wu2020tree} & C3D  & 37.39 & 17.69 & 37.22 \\ \hline
\multirow{3}{*}{PB} & 2D-TAN \cite{zhang2020learning} & VGG & 39.70 & 23.31 & - \\
 & MMN \cite{wang2022negative} & VGG & 47.31 & 27.28 & -  \\
 & FVMR \cite{gao2021fast} & I3D & 55.01 & 33.74 & -  \\
 & MS-2D-TAN \cite{zhang2021multi} & I3D & 56.64 & 36.21 & -  \\ 
 \midrule
\multirow{6}{*}{PF} 
 & VLSNet \cite{zhang2020span} & C3D & 47.31 & 30.19 & 45.15 \\
 & LGI \cite{mun2020local} & I3D & 59.46 & 35.48 & 51.38 \\
 & HISA \cite{xu2022hisa} & I3D  & 61.10 & 39.70 & 53.57 \\ 
 & UnLoc \cite{yan2023unloc} & CLIP  & 60.80 & 38.40 & - \\ 
 & UniVTG \cite{lin2023univtg} & SF+CLIP  & 60.19 & 38.55 & 52.17 \\ 
 \cmidrule{2-6}
 & Moment-DETR \cite{lei2021detecting} & SF+CLIP  &  53.63 & 31.37 & - \\
 & $\text{Moment-DETR}^\dag$ \cite{lei2021detecting} & SF+CLIP  & 53.23 & 31.21 & 46.74 \\
 & \textbf{Moment-DETR+Ours} \cite{lei2021detecting} & SF+CLIP  & \textbf{56.85} & \textbf{32.96} & \textbf{48.90} \\
 \cmidrule{2-6}
 & QD-DETR \cite{moon2023query} & I3D & 50.67 & 31.02 & - \\
 & $\text{QD-DETR}^\dag$ \cite{moon2023query} & I3D & 59.22 & 36.72 & 50.50 \\
 & \textbf{QD-DETR+Ours} \cite{moon2023query} & I3D & \textbf{61.72} & \textbf{40.48} & \textbf{52.55} \\
\bottomrule

\end{tabular}
}
\end{table}

\subsection{More Qualitative Results}

In this section, we provide more visual examples to demonstrate the effectiveness and superiority of our method. 

\subsubsection{Visualizations of Saliency Scores}
\begin{figure}[th]
  \centering
  \begin{subfigure}{0.48\linewidth}
    \includegraphics[width=\textwidth]{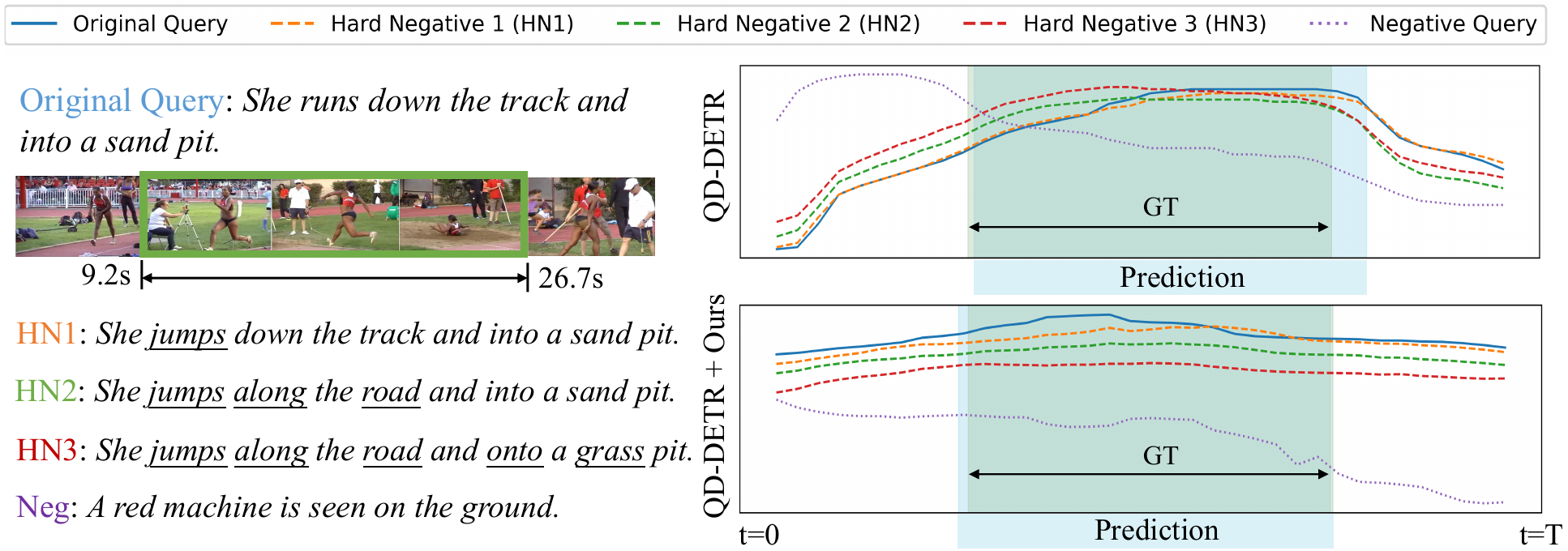}
    \caption{}  % v\_bZ4r3Y\_qceE.mp4
    % \vspace{3pt}
    \label{fig:v_bZ4}
  \end{subfigure}
  % \hspace{5pt}
  \begin{subfigure}{0.48\linewidth}
    \includegraphics[width=\textwidth]{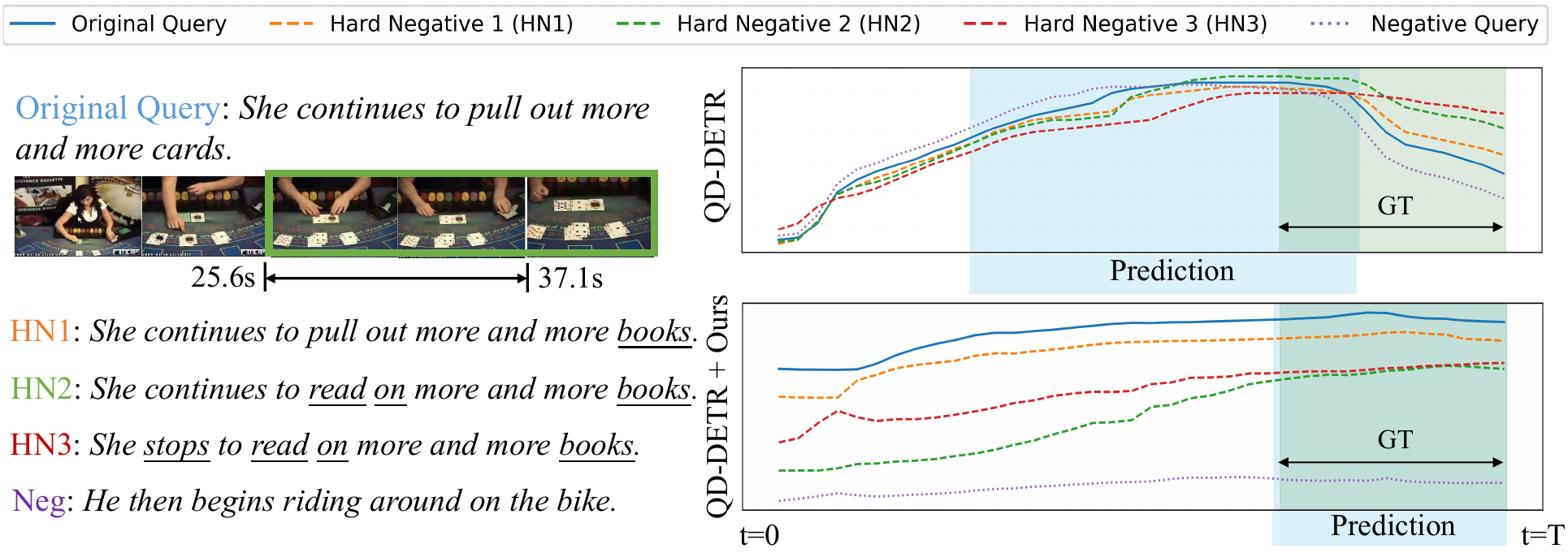}
    \caption{}  % v\_xsBFnpdLWkU.mp4
    % \vspace{3pt}
    \label{fig:fail}
  \end{subfigure}
  \caption{Visualization of saliency scores given different queries in ActivityNet-CG.}
  \label{fig:anet-saliency}
\end{figure}

First, we provide more visualization of the saliency scores in ActivityNet-CG. We observe that the existing work has difficulty recognizing hard negative queries, showing irrational saliency responses. For instance, in \cref{fig:v_bZ4}, the hard negative query “\textit{She \underline{jumps along} the \underline{road} and \underline{onto} a grass pit}” is even more salient than the positive query “She runs down the track and into a sand pit”. The irrational responses lead to unprecise moment localization since there is no corresponding moment of this negative query in the video. In contrast, our approach consistently improves the model’s ability to distinguish between different words in positive and hard negative queries and yield hierarchical responses, thereby achieving better moment localization and compositional generalization. 

\begin{figure}[tb]
  \centering
  \begin{subfigure}{\linewidth}
    \includegraphics[width=\textwidth]{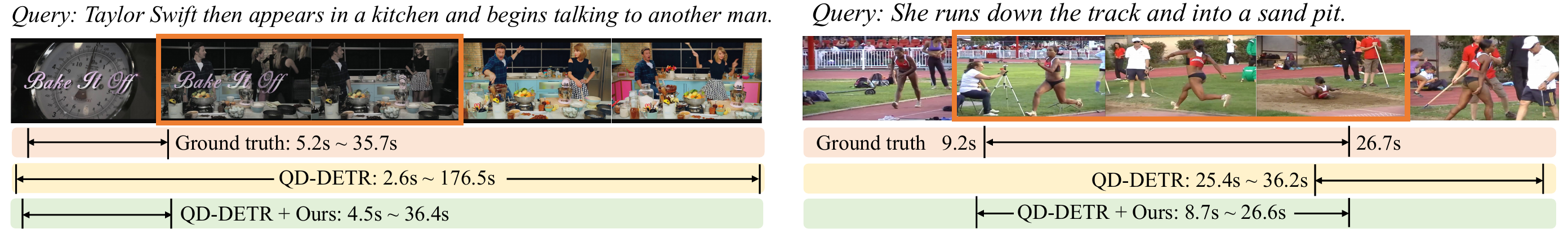}
    \caption{Two Samples from the Test-Trivial split}
    \vspace{5pt}
    \label{fig:anet-tt}
  \end{subfigure}
  % \vspace{5pt}
  \begin{subfigure}{\linewidth}
    \includegraphics[width=\textwidth]{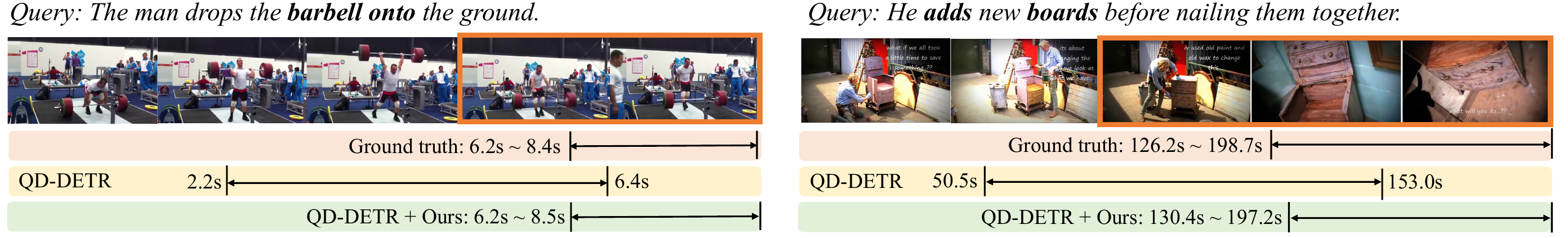}
    \caption{Two Samples from the Novel-Composition split}
    \vspace{4pt}
    \label{fig:anet-nc}
  \end{subfigure}
  \begin{subfigure}{\linewidth}
    \includegraphics[width=\textwidth]{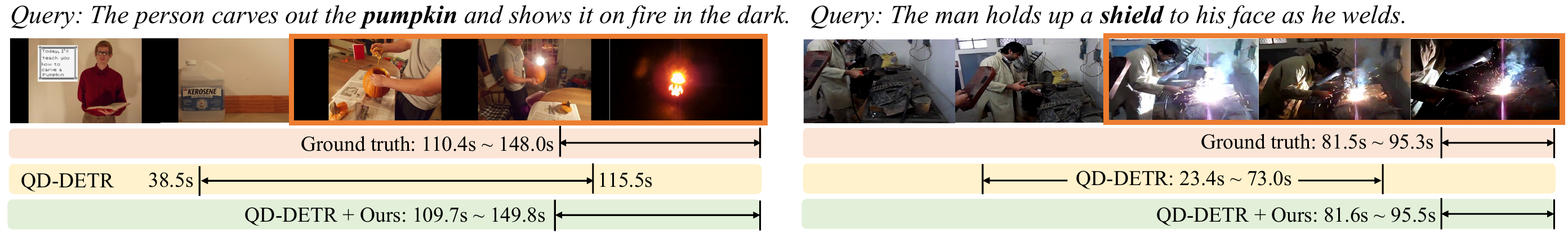}
    \caption{Two Samples from the Novel-Word split}
    \label{fig:anet-nw}
  \end{subfigure}
  \caption{Qualitative comparisons between QD-DETR and QD-DETR+Ours on samples from different test splits of ActivityNet-CG.}
  \label{fig:anet-vis}
\end{figure}

% Existing work has difficulty in recognizing hard negative queries, showing irrational saliency responses. Our approach consistently improves the model's ability to distinguish between different primitives in a query sentence and achieves better compositional generalization.

However, in the failure case \cref{fig:fail}, we find that our method sometimes fails to distinguish the subtle differences in the Hard Negative 2 and 3 but still responds rationally to the Positive, Hard Negative 1, and the Negative queries. We assume that our method struggles with longer videos and sentences in ActivitiNet-CG.

In these saliency score visualizations, we find that the gap of saliency scores derived from positive and negative queries of our method is more discriminative than that of the baseline model. We suggest that it is the inter loss $\mathcal{L}_{inter}$ in the coarse-grained ranking loss $\mathcal{L}_{cr}$ that plays a role in improving the model's ability to discriminate between positive and negative samples. Moreover, the magnitude of the saliency score within the GT is less significant than that of the baseline. We assume this is due to the larger gap between the positive and negative samples leading to less pronounced saliency score variation in the positive sample. However, this does not impair the model's ability to localize moments of positive samples accurately.

\subsubsection{Visualizations of Temporal Grounding}
We offer more visualization results of the moment predictions in ActivityNet-CG in \cref{fig:anet-vis}. Our method can enhance the existing work to generalize to Novel-Composition and Novel-Word testing as well as predict the moment more precisely in the IID Test-Trivial split.

In the examples in ActivityNet-CG, our approach shows its adaptability to longer videos and sentences. Despite the queries in \cref{fig:anet-tt} “Taylor Swift then appears in a kitchen and begins talking to another man” and "She runs down the track and into a sand pit" only containing known words and compositions, the baseline inaccurately identifies the relevant moment, whereas our method exhibits a more precise alignment with the ground truth. For the Novel-Composition queries (\cref{fig:anet-nc}) "\textit{The man drops the \textbf{barbell onto} the ground}" and "\textit{He \textbf{adds} new \textbf{boards} before nailing them together}", our approach delivers better temporal accuracy, closely reflecting the ground truth. Upon encountering Novel-Word queries such as "\textit{The person carves out the \textbf{pumpkin} and shows it on the fire in the dark}" and "\textit{The man holds up a \textbf{shield} to his face as he welds}" in \cref{fig:anet-nw}, our method proficiently generalizes to the unseen primitives "\textit{\textbf{pumpkin}}" and "\textit{\textbf{shield}}".

% The visualization results demonstrate that our approach successfully directs DETR-based models to leverage hierarchical negative samples, thereby improving their ability to generalize to unseen compositions and words.

\section{Limitations and Future Work}
Existing large language models are highly sensitive to prompt templates, and the negative queries generated by different templates may vary in effectiveness, \eg, we noticed that some negative queries generated by the LLM still lack semantic feasibility and may include words that do not exist in the dictionary. It is worth exploring how to design effective prompt templates as well as data curation strategies to produce better negative queries. In addition, we only consider novel compositions in the query sentence without taking visual-level compositions into account, which might facilitate the creation of better compositional vision-language representations.

\end{document}